\documentclass[default,iicol]{sn-jnl}

\jyear{2022}%

\theoremstyle{thmstyleone}%
\theoremstyle{thmstyletwo}%

\theoremstyle{thmstylethree}%
\usepackage{color}
\usepackage{url} 
\definecolor{GREY}{RGB}{170,170,170}
\begin{document}

\title[Article Title]{Towards Language-guided Visual Recognition via \\Dynamic Convolutions }


\author[1]{\fnm{Gen} \sur{Luo}}\email{luogen@stu.xmu.edu.cn}

\author*[1,2]{\fnm{Yiyi} \sur{Zhou}}\email{zhouyiyi@xmu.edu.cn}

\author[1,2]{\fnm{Xiaoshuai} \sur{Sun}}\email{\{zhouyiyi, xssun,rrji\}@xmu.edu.cn} 


\author[3]{\fnm{Yongjian} \sur{Wu}}\email{littlekenwu@tencent.com}


\author[4]{\fnm{Yue} \sur{Gao}}\email{gaoyue@tsinghua.edu.cn} 

\author[1,2]{\\ \fnm{Rongrong} \sur{Ji}}

\affil[1]{Key Laboratory of Multimedia Trusted Perception and Efficient Computing,  
\\ Ministry of Education of China, Xiamen University, 361005, P.R. China.}

\affil[2]{Institute of Artificial Intelligence, Xiamen University, 361005, China}

\affil[3]{Youtu Lab, Tencent}

\affil[4]{Software School of Tsinghua University}


\abstract{		In this paper, we are committed to establishing a unified and end-to-end  multi-modal  network via exploring   language-guided visual recognition. 
	To approach this target, we first propose a novel multi-modal convolution module called \textit{Language-guided Dynamic Convolution} (LaConv). Its convolution kernels are dynamically generated based on natural language information, which can help extract differentiated visual features for different multi-modal examples.
	Based on the LaConv module, we further build a fully  language-driven convolution network, termed as {\textit{LaConvNet}}, which can unify the visual recognition and multi-modal reasoning in one  forward  structure. To validate  {LaConv} and LaConvNet, we conduct extensive experiments on seven benchmark datasets of three vision-and-language tasks, \emph{i.e.}, visual question answering (VQA), referring expression comprehension (REC) and segmentation (RES).  The experimental results not only show the  competitive  or better performance of LaConvNet  against   existing multi-modal networks, but also witness the merits of LaConvNet as an unified structure, including compact network,  low computational cost and  high generalization ability. Our source code is released in SimREC project: \url{https://github.com/luogen1996/LaConvNet}. }

\keywords{Referring Expression Comprehension, Visual Reasoning, Vision and Language}



\maketitle

\section{Introduction}


In recent years, the rapid development of   joint vision-language study have been supported by a flurry of benchmark datasets~\cite{liu2019clevr,antol2015vqa,goyal2017making,johnson2017clevr,hu2017modeling,kazemzadeh2014referitgame,REFCOCOG,hudson2019gqa,perez2018film} and methods~\cite{anderson2018bottom,hudson2018compositional,hudson2019learning}. The latest research trend~\cite{perez2018film,hudson2018compositional,santoro2017simple,mascharka2018transparency,mao2019neuro} has gone beyond a simple understanding of multi-modal information~\cite{morency2011towards,zadeh2016mosi,poria2015deep}, and focused on more advanced cognitive tasks, such as visual reasoning~\cite{zhang2020text,johnson2017clevr,liu2019clevr,hudson2019gqa}, visual question answering (VQA)~\cite{antol2015vqa,goyal2017making,johnson2017clevr,hudson2019gqa,krishna2017visual,zhou2019plenty}, and referring expression comprehension  (REC)~\cite{plummer2015flickr30k,liu2019clevr,hu2017modeling,kazemzadeh2014referitgame,REFCOCOG}.

\begin{figure}[t]
	\centering
	\includegraphics[width=1\columnwidth]{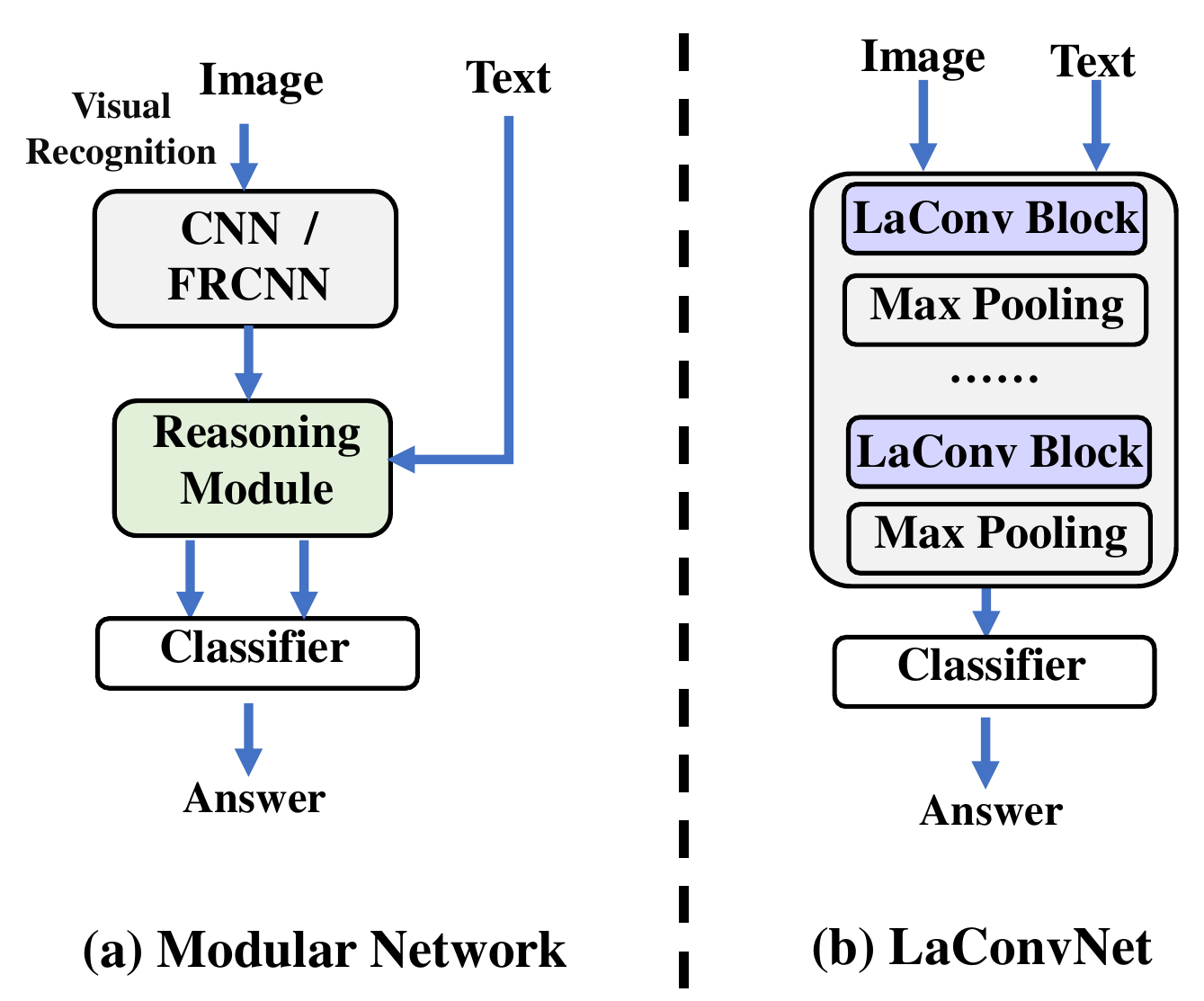}
	
	\caption{Comparison between the conventional modular network and  our unified model. 
		 The proposed LaConv  block can  combine   visual recognition and multi-modal interaction into one processing step. Based on it, we   build a unified and end-to-end network called LaConvNet. } 
	\label{fig1}  
	\vspace{-1em}
\end{figure}

%
%

To accomplish these tasks, most existing vision-and-language  (VL) systems adopt a modular structure.    As shown in Fig.~\ref{fig1} (a), a typical VL system often uses a  visual backbone, \emph{e.g.}, ResNet~\cite{he2016deep} or FasterRCNN~\cite{ren2015faster}, to extract the features of the input image, based on which another inference network is deployed to model   cross-modal interactions.  
This long-popular   paradigm has achieved great success in various VL tasks~\cite{antol2015vqa,goyal2017making,johnson2017clevr,hu2017modeling}, but has been also criticized for its excessive parameters and  high computational overhead~\cite{kim2021vilt,xu2021e2e,simvlm}.

In contrast to this modular structure,  we are committed to establishing an  unified alternative   by  exploring  language-guided visual recognition  from raw pixels. 
As shown in Fig.~\ref{fig1} (b), we aim at embedding   language information into the process of visual recognition,  and then directly output the language-dependent  visual features. 
This motivation is inspired by the human cognitive mechanism towards   multi-modal tasks. 
Neuroscience researches~\cite{stein2008multisensory,mcgurk1976hearing,shams2002visual,bonath2007neural}  show that  \textit{primary visual cortex cells} can be  influenced by other modalities, \emph{e.g.}, text or sound, and then produce     multimodal sensory. It means that    low-level visual recognition can also be driven by   natural language signals.
For example, after receiving a natural language instruction, people will  perform the visual recognition related to the instruction, \emph{e.g.}, paying attention to relevant regions and analyzing    information like colors or textures. 

To achieve this target, we first propose a novel \emph{Language-Guided Dynamic Convolution} (LaConv) module, of which structure is depicted in Fig.~\ref{arch}. 
The property of ``\textit{language-guided visual recognition}'' are mainly reflected in that LaConv can realize differentiated feature extractions on the same image according to different natural language instructions. This property is attributed to its dynamic convolution filters predicted by natural language information. 
Through this novel multi-modal convolution, LaConv can complete  visual recognition and multi-modal reasoning in one processing step.

Based on LaConv, We build  a fully language-guided convolution network, termed \emph{LaConvNet}.  As shown in Fig.\ref{fig1},  LaConvNet processes the input image from the pixel level and embeds the natural language information into the  whole  process  of  visual feature learning.  The output visual features can be directly used for multi-modal prediction.  Such a property makes it highly generalizable to different VL tasks. For example, we can simply add a detection layer after LaConvNet for REC task.  Compared to previous multi-modal networks,  LaConvNet gets rid of   large  image backbones and unifies  feature extraction and    multi-modal inference  in one  forward structure.

To validate our approach, we  first conduct   experiments on four referring expression comprehension (REC) benchmark datasets, \emph{i.e.},    RefCOCO~\cite{nagaraja2016modeling}, RefCOCO+~\cite{nagaraja2016modeling}, RefCOCOg~\cite{REFCOCOG}, Referit~\cite{kazemzadeh2014referitgame} and Flickr30k~\cite{plummer2015flickr30k}.  After pre-training on a  certain amount of  data, \emph{e.g.,} Visual Genome~\cite{krishna2017visual}, LaConvNet is comparable to a set of state-of-the-art (SOTA) methods~\cite{Luo_2020_CVPR,yang2020improving,transvg} and large-scale pre-training models~\cite{uniter,ernie},  while maintaining superior  compactness and efficiency.  In addition, we validate  its  generalization ability  on two visual reasoning datasets, \emph{i.e.,} CLEVR~\cite{johnson2017clevr} and CLEVR-Ref+~\cite{liu2019clevr}, where   competitive performance can be also witnessed. 

 In  summary, our contributions are three-fold: 
\begin{itemize}
	\item We proposed an efficient language-guided dynamic convolution block, namely \emph{LaConv}, which can simultaneously   accomplish visual feature learning and  multi-modal inference.
	\item  Based on LaConv, we proposed LaConvNet,  a compact, efficient and highly generalized network for language-guided visual recognition.
	\item On multiple benchmarks, LaConvNet is on par with or even better than existing modular methods, while retraining superior model compactness and efficiency.
\end{itemize}

\section{Related Work}
\subsection{Referring Expression Comprehension and Segmentation}
Referring expression comprehension (REC) and Segmentation (RES)~\cite{kazemzadeh2014referitgame,REFCOCOG,plummer2015flickr30k} are two popular tasks in vision-and-language  research, which  aim  to locate the target object according to a given natural language expression.   In particular, REC and RES locate the referent by bounding box and object mask, respectively. Early REC and RES works ~\cite{mattnet,ddpn,cmatt,nmtree} usually follow the two-stage pipeline.   They  typically use  Faster-RCNN~\cite{ren2015faster} to extract  visual representations of salient objects, and then compare them to the language features to select the  best-matching  one.  Due to the   multi-step  setup,  these approaches are often  inferior in their inference speed and performance.  To this end, recent researchers have  turned  their attention to one-stage   modeling~\cite{yang2019fast,realgin,transvg,yang2020improving,Luo_2020_CVPR,lbyl,itershrinking,VLT,LAVT,CGAN,refclip,refteacher}.   Specifically, these approaches first  use convolution neural networks like ResNet~\cite{he2016deep} or DarkNet~\cite{yolov3} as the visual backbone, based on which a multi-modal branch is deployed for cross-modal interactions.  Then, a detection or segmentation head  is used to directly output the prediction of the referent.  Inspired by the great success of Transformer~\cite{vaswani2017attention},  recent advances focus on applying Transformer in REC~\cite{transvg,lwtransformer,trar}, and RES~\cite{VLT,LAVT,CMSA,ReSTR} to improve the detection ability and the reasoning ability.

Despite of the effectiveness, most approaches   still rely on   heavy visual backbones and expensive multi-modal fusion blocks.   By contrast, LaConvNets  simplify this pipeline with language-guided visual recognization, making them  much more  efficient than existing REC and RES approaches.

\subsection{Visual Reasoning}
Visual reasoning is  an important research direction in vision and language study, which aims to examine the compositional, relational and structural reasoning abilities of multi-modal models.  
Existing visual reasoning  methods can be categorized into two groups: the neural-based~\cite{hudson2018compositional,yang2016stacked,santoro2017simple,perez2018film} and the symbol-based approaches~\cite{johnson2017inferring,hu2018explainable,mascharka2018transparency,yi2018neural,mao2019neuro,hu2017learning,andreas2016neural}, respectively. In particular,  neural-based approaches~\cite{hudson2018compositional,yang2016stacked,santoro2017simple,perez2018film}   aim to reason over the image based on the language instructions via attention~\cite{hudson2018compositional,yang2016stacked}, relational network~\cite{santoro2017simple} or modulated network~\cite{perez2018film,santoro2017simple}.  
Despite of the effectiveness,  they  are often inferior in their interpretability and training efficiency.   In this case, some works~\cite{johnson2017inferring,hu2018explainable,mascharka2018transparency,yi2018neural,mao2019neuro,hu2017learning,andreas2016neural} explore symbol-based networks to decompose visual reasoning into several pre-defined programs.    These models usually learn a symbolic predictor to parse the language instructions into specific programs, and then execute the corresponding  modules to predict the final answer.    This methodology can  ensure  the   interpretability and efficiency of reasoning process, but  the models still heavily rely  on expert  knowledge in program design and  require  additional program annotations.

  Compared to these approaches,   LaConvNet mainly differs  in its unified multi-modal structure, which can simultaneously perform    visual recognition and  multi-modal reasoning from raw pixels.  Such a structure make LaConvNet more compact and efficient than existing approaches while still maintaining strong reasoning capabilities.  In addition, we also show that as a neural-based approach, LaConvNet is still  interpretable.

\subsection{Language-guided  Convolutions}

Applying convolution to  process vision and language data is not new in literature. In early attempts, convolution is used as a cross-modal module to jointly process image and text features~\cite{perez2018film,nguyen2020revisiting,de2017modulating,gao2018question}. 
In terms of language-driven convolution, there are several recent methods related to our work.  
 Specifically, FiLM~\cite{perez2018film} aims to learn normalization weights by  language features and applies them after convolutions.  With this design,  FiLM can extract language-related visual features and achieve multi-modal reasoning. 
 Gao \emph{et al.}~\cite{gao2018question}  propose a question-guided hybrid convolution (QGHC) module, of which the convolution kernels are  also predicted by the text features.  However, these dynamic kernels are  independent of  images and may not adapt well to different visual content. Qiu \emph{et al.}~\cite{qiu2020language} propose language-aware deformable convolution  to learn  fine-grained multi-modal representation. In  \cite{qiu2020language}, the offsets of the deformable convolution are predicted by   multi-modal features. 
   Compared to these approaches, LaConv  can generate different language-driven convolution kernels for different visual areas, therefore  achieving better vision-and-language alignments.   In addition, all existing multi-modal convolutions are used as plug-in components for existing modular networks. In this paper, we use the propose LaConv as a stand alone building block for fully language driven convolution network.    In  practice, using previous convolution modules to build the unified network  are also inferior  to LaConv.

\subsection{Efficient and Unified Multi-modal Network}
Recently, the design of unified multi-modal networks has gained gradually increasing interest in Vision-Language (VL) community~\cite{kim2021vilt,simvlm,uniter,villa,blip2}.  However, most existing   multi-modal models~\cite{uniter,villa,blip2} still adopt the highly redundant architecture, which consists of a large visual backbone and deep multi-modal fusion branches. This computational expensive architecture greatly increases the costs of deployment. To improve the efficiency, some works~\cite{kim2021vilt,simvlm}  handle visual processing and multi-modal fusion into a single unified manner.  Among them, the most representative work is ViLT~\cite{kim2021vilt} which directly    feeds image  and text embedding into a  vision transformer~\cite{vit} to obtain joint vision-and-language representations. After pre-training on millions of image-text pairs, ViLT can be applied to various VL tasks, \emph{e.g.,} VQA. Recently, Wang et.al~\cite{simvlm} propose a ViT-like model, which is pre-trained   with a novel  pre-training objective, \emph{i.e.,} prefix language modeling. Compared to ViLT, it further insert  more convolution blocks at the beginning of the network. 

Recently, there are also several methods~\cite{Unit,unified-io} proposed to explore the unified learning of various VL tasks. However, these methods focus more on the multi-task learning by using one multi-modal network, and their network structures are still modular and still requires large visual backbones. In this case, the contributions of these works are orthogonal to this paper.

Despite of the great success,  these Transformer based models are still over-parameterized and computationally expensive.  In this case, we explore an extremely  compact unified multi-modal network called LaConvNet.   Due to  the efficient design of language-guided dynamic convolution, LaConvNet can be much more  efficient than existing unified multi-modal models.

\section{ Language-guided Dynamic Convolution }
In this section, we give the definition of the proposed  Language-guided Dynamic Convolution (LaConv), of which structure is illustrated in Fig.\ref{fig1}. 
We first introduce   its main principle  of  language-guided  dynamic convolution, and then describe how its convolution kernels are generated from text features.

\begin{figure*}[t]
	\centering
	\includegraphics[width=2\columnwidth]{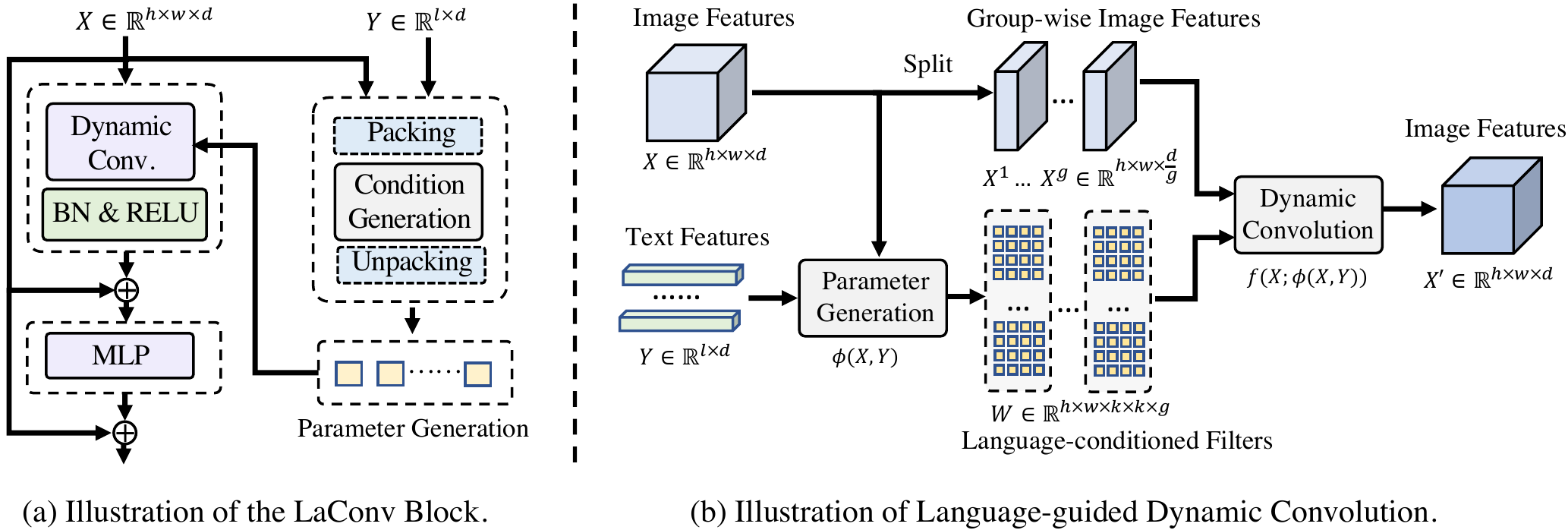}
	
	\caption{Illustration of the  LaConv block and its language-guided dynamic convolution.  (a). In  LaConv block, the spatial-wise convolution kernels are generated by the language conditions, which are then used to dynamically process the visual features.   (b). The parameter generation module firstly produces language-conditioned convolution filters based on image and text features.  After that, language-guided dynamic convolution is conducted to process the image features. }
	\label{dyconv}
\end{figure*}

\subsection{Definition.}
\label{sec:dy_conv}
To achieve language-guided visual recognition, we propose a LaConv module in this paper. Compared to existing multi-modal modules, such as self-attention~\cite{VLT,CMSA,vlt_cvpr}, the main difference of LaConv is that it can produce language-conditioned dynamic kernels for depth-wise convolution, thereby achieving differentiated visual feature learning for different input texts. 

Specifically, given the image features $X \in \mathbb{R}^{h \times w \times d}$ and the text features $Y \in \mathbb{R}^{l \times d}$, LaConv is defined by 

\begin{equation}
\begin{aligned}
X'=f(X;\phi(X,Y;\theta)).
\end{aligned}
\label{dyconv_1}
\end{equation}
Here,  $f(\cdot)$ denotes the convolution operation, $\phi: X,Y \rightarrow W$ is the   function of parameter generation based on image and text features, and $\theta$ denotes the learnale parameters.  Compared with common convolutions~\cite{he2016deep}, of which kernels are static for all images, LaConv can produce dynamic convolution kernels based on the input text, thereby achieving  language-guided visual recognition.

 To  improve the efficiency of LaConv, $f(\cdot)$  and $\phi(\cdot)$  are designed as  an efficient operator and   a lightweight parametric function, respectively.
Specifically,  we define $f(\cdot)$  as a  dynamic depth-wise convolution.  Given the language-conditioned convolution kernel $W \in \mathbb{R}^{h \times w \times k \times k \times g}$, $f(\cdot)$ can be formulated by 

\begin{equation}
\small
\begin{aligned}
\text{dyconv}(X^l)_{i,j}&= \sum_{\Delta i=1}^{k}\sum_{\Delta j=1}^{k} {X}^l_{ i+\Delta i, j+\Delta j} \odot {W}_{ i,j, \Delta i,\Delta j,l}, \\
f(X;W)&=[\text{dyconv}(X^1),...,\text{dyconv}(X^{g})].
\end{aligned}
\label{dyconv_2}
\end{equation}
Here, $k$ and $g$  are the kernel size and the number of groups.  $[\cdot]$ denotes the  concatenation.  In practice, LaConv is much more efficient than common convolutions.  Specifically, the FLOPs of $f(\cdot)$ are $\frac{1}{d}$ of common convolutions, where $d$ is usually up to hundreds.   Notably,  although the  language-conditioned convolution kernel $W$ has a large shape, it can be predicted via $\phi(\cdot)$ with lightweight parameters. We discuss the parameter generation  in next section.

From Eq.~\ref{dyconv_2},  we can also observe two key differences of LaConv from the existing depth-wise convolution~\cite{krizhevsky2012imagenet}.   Firstly,  in LaConv, each image position $(i,j)$ has their corresponding filters, which is to model the spatial  relationships in text information, \emph{e.g.}, ``left person''. 
When filters are shared, such information is hard to recognize. 
Secondly, with the language-conditioned convolution filters $W$, LaConv can extract differentiated visual features   based on different texts.  
 
LaConv also has a distinct design principle different from existing multi-modal modules~\cite{VLT,LAVT,CGAN,transvg} that mainly apply self-attention for cross-modal interaction. The applied dynamic convolution can help LaConv achieve better visual feature learning, and the used language-conditioned parameter generation can also facilitate the cross-modal interactions for various VL tasks. Meanwhile, its computation efficiency is also superior than the attention-based modules.

\begin{figure*}[t]
	\centering
	\includegraphics[width=2\columnwidth]{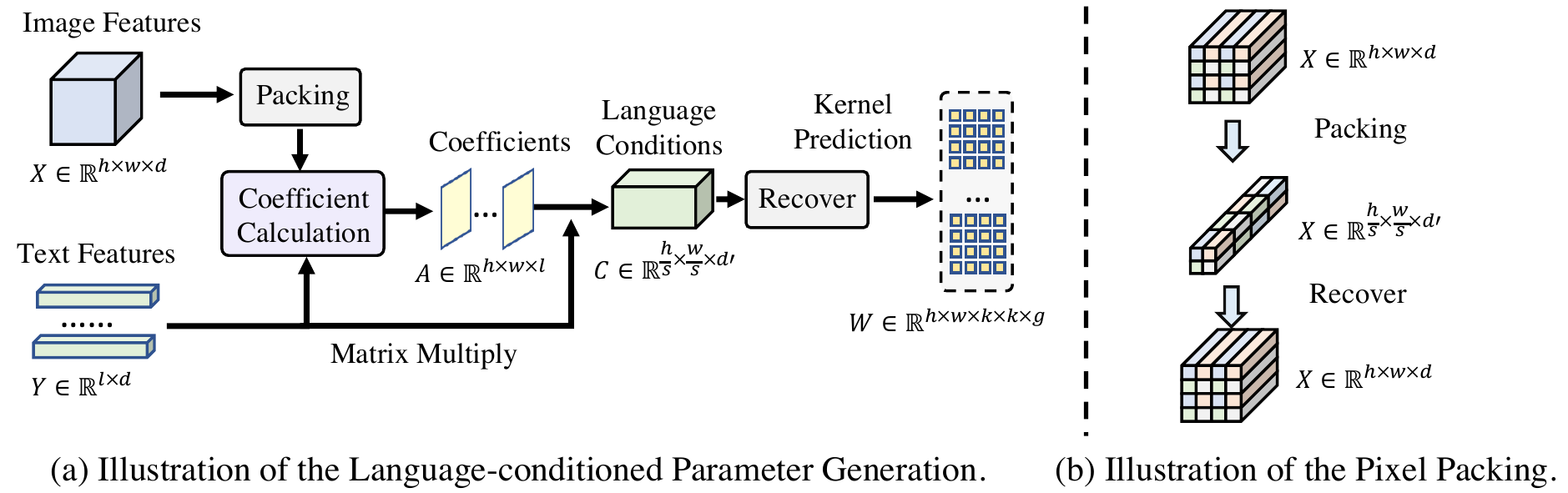}
	
	\caption{Illustration of the parameter generation  and pixel packing. The  input text features  are first transformed into a language condition matrix, which is spatially and semantically related to the image features. 
		Then, the convolution filters   are predicted based on this condition matrix.
		 \emph{Pixel packing} operations are  applied to alleviate the \emph{low-rank degeneration} in low-level visual features. }
	\label{condgen}
\end{figure*}

\subsection{Language-conditioned Parameter Generation}\label{sec:lcpg}
From the definition  of LaConv,   the key to realize language-guided convolution lies in the    design of the  function of parameter generation, \emph{i.e.,} $\phi(\cdot)$.
As shown in Fig.~\ref{condgen}, $\phi(\cdot)$ needs to generate dynamic convolution filters based on  natural  language information. These filters  are  spatially and semantically related to the image.  In this case, the processing of LaConv can be dynamically adjusted based on the changes of visual content and text information.

Generally,   the length of the  text features is often not consistent with the resolution of the image features, and they are also not spatially aligned. 
To this end, we first  transform the text features into a condition matrix ${C} \in \mathbb{R}^{(h\times  w) \times d}$, which has the same shape as the image ones.  Each   condition vector of $C$ is also semantically related to the corresponding image region.  In this case,  the parameter generation function $\phi(\cdot)$ can be defined by 

\begin{equation}
\begin{aligned}
\phi(X,Y;\theta)&= C{W}_1+b_1,\\
\text{where} \quad {C} &= \sigma({A} ({Y}{W}_A){W}_C).
\end{aligned}
\label{eq3}
\end{equation}
%
Here, $W_1 \in \mathbb{R}^{d \times (k\times k \times g)}$  and  $b_1 \in \mathbb{R}^{(k\times k \times g)}$ are the projection weight and bias term for predicting convolution kernels.  $W_A \in \mathbb{R}^{d \times d }$ and $W_C \in \mathbb{R}^{ d \times d}$ are two projection weights for the generation of  condition matrix.  
${A}\in \mathbb{R}^{\left(h\times w\right)\times l}$ is the affinity matrix between $X$ and $Y$, and its values denotes the coefficients between the features of two modalities. Here, we resort to  \emph{scaled dot-product attention}~\cite{vaswani2017attention} to compute the multi-modal coefficients:

\begin{equation}
\begin{aligned}
{A}=\text{Softmax}(\frac{(XW_X)^T (YW_Y)}{\sqrt{d}}),
\end{aligned}
\label{att_func}
\end{equation}
where $W_X \in \mathbb{R}^{d \times d }$ and $W_Y \in \mathbb{R}^{d \times d}$ are two projection weights.
To improve   module capacity, we also  extend Eq.~\ref{att_func}   into a \emph{multi-head} version~\cite{vaswani2017attention}.  The mechanism of multi-modal coefficient generation is related to self-attention~\cite{vaswani2017attention}, but we only conduct attentions between two modalities, which is practically computational friendly.

With Eq.\ref{eq3} and Eq.\ref{att_func}, the obtained  condition matrix not only has the same shape as the visual features, but also  spatially relates  to the image regions. 
Therefore, through the condition matrix, we can effectively embed language information into   dynamic  parameter generations  for each visual region  and   multi-modal example.  Meanwhile,   our parameter generation is  actually more  parameter-efficient than conventional convolutions, and   its parameters are  about $\frac{3}{k^2}$  of the common  convolution layer.

 	

\begin{figure*}[t]
	\centering
	\includegraphics[width=2\columnwidth]{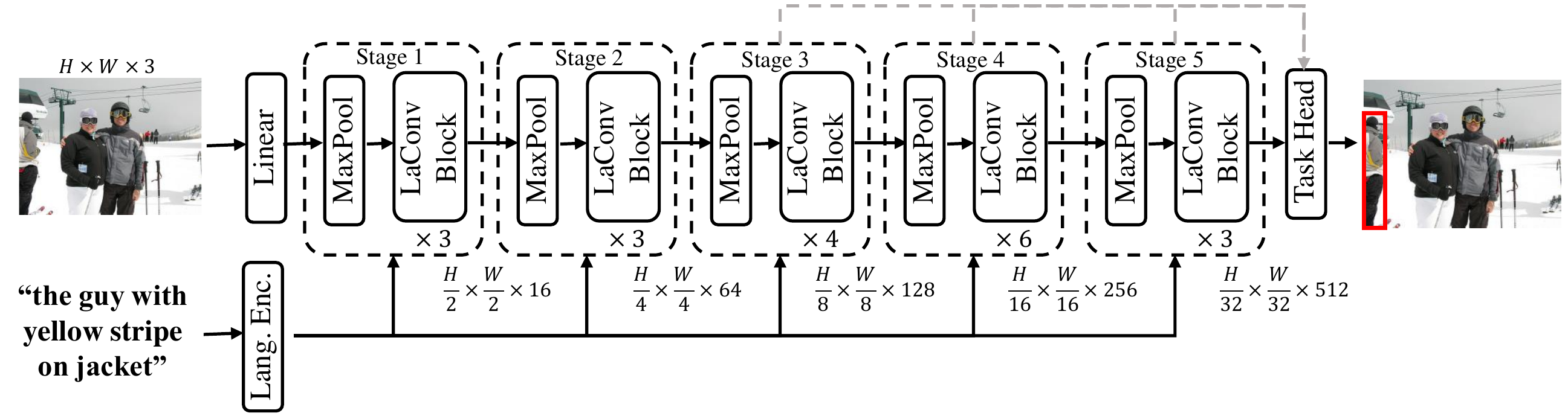}
	
	\caption{ The architecture of LaConvNet. LaConvNet consists of 5 stages, each of which has a max pooling layer and several  LaConv blocks.  The input image and text are firstly processed by a linear projection and a language encoder, respectively. Afterwards, LaConv blocks are used to exact visual features based on language information. Finally, the task-specific head is applied to predict the results of the task.    } 
	\label{arch} 
	\vspace{-1em}
\end{figure*}

\label{pixelpacking}
\textbf{Pixel packing.} Since the generation of   condition matrix is based on the scale-dot product between two types of features, it  is prone to the issue of \emph{low-rank degeneration}~\cite{bhojanapalli2020low} when the number of the image features is much larger than  their feature dimension,  \emph{e.g.}, the low-level image features.

To this end, we    introduce a tensor operation called  \textit{pixel packing} to alleviate this problem. 
As shown in Fig~\ref{condgen}, given the image  features $\textbf{I}\in \mathbb{R}^{\left(h\times w\right) \times d}$, we first pack them into $k$ new visual tokens, where $k=(h\times w)/s^2$, and each token is the concatenation  of the  $s\times s$ local features. 
So, the number of new image features is reduced by $s^2$ times. Correspondingly, the resolutions of the condition matrix ${C}$ and the affinity matrix ${A}$ in Eq.~\ref{eq3} also becomes $(h\times w)/s^2$. 
Before the parameter predictions, we will recover the condition matrix back to the size of $(h\times w)$, which is to make the predicted filters to adapt the input image features.  
 
Overall, pixel packing is a  packing-and-recover processing,  which can alleviate the issue of low-rank degeneration via reducing the feature resolutions, while maintaining the multi-modal interactions.  In this case, its mechanism and effect are still different to other  tensor operations like patch embedding~\cite{vit}.

\section{LaConvNet\label{net_intro}}
Based on LaConv, we further propose a unified and end-to-end   network, termed \emph{LaConvNet}, as shown in Fig.~\ref{arch}.   LaConvNet  processes  images directly from  the pixel level, completely abandoning the traditional convolution backbones like ResNet~\cite{he2016deep} or MaskRCNN~\cite{he2017mask}.
This property is the main difference  between LaConvNet and  most existing VL systems, which also makes LaConvNet much more compact.   
Specifically, we first construct the basic LaConv block, and the $i$-th layer  can be formulated by

\begin{equation}
\label{equ:laconv_block}
\begin{aligned}
X_i'&=\sigma \left(\text{BN}\left(f\left(X_{i-1};\phi\left(X_{i-1},Y\right)\right)\right)+X_{i-1}\right),\\
X_i&=\text{MLP}(X_i')+X_i',
\end{aligned}
\end{equation}
where $\sigma$ denote the activation function and BN is the batch normalization layer. In Eq.~\ref{equ:laconv_block}, the MLP layer is formulated by

\begin{equation}
\label{equ:mlp}
\begin{aligned}
\text{MLP}(X)=\text{BN}\left(\sigma\left(\text{BN}\left(XW_a\right)\right)W_b\right), 
\end{aligned}
\end{equation} 
where $W_a \in \mathbb{R}^{d \times 4d}$ and $W_b \in \mathbb{R}^{4d \times d}$ are two projection weights.

Based on the LaConv block,  we propose  two  versions of LaConvNet,  termed \textit{{LaConvNet-S}}  and \textit{LaConvNet-B}.  LaConvNet-S and LaconvNet-B contain  10 and 19 Laconv blocks, respectively.    For each stage of LaConvNet, we set a proper \textit{packing size } $s$ to keep the expressive power of the parameter generation.  The detailed   configurations are shown in Tab.~\ref{net}.

Notably, the design of {LaConvNet} still has a large space for exploration, such as the choice of depth, width, filter size \emph{etc.}
In this paper, we only aim to provide an effective baseline network  to quickly validate our  motivation.

\begin{table*}[t] 
			\centering
	\caption{ Network architecture of {LaConvNet}. ``$n-d$'' denotes the  channels of  transformations.  ``s'' denotes the packing size of the LaConv block. Similar to ResNet, {LaConvNet} contains 5 stages and each stage has several {LaConv} blocks.  }
 \setlength\tabcolsep{25pt}
	\begin{tabular}{cccc}
		\toprule
		\multicolumn{1}{c}{Output Size}      & s          & LaConvNet-S                                                                                                                & LaConvNet-B                                                                                                                 \\ \midrule
		\multicolumn{1}{c}{$224 \times 224$} & -                     & 16-d linear                                                                                                                 & 16-d linear                                                                                                                   \\ [4.5pt]
		\multicolumn{1}{c}{$112 \times 112$} & -                     & 2x2, stride 2 pool                                                                                                                & 2x2, stride 2 pool                                                                                                       \\ [4.5pt]
		\multicolumn{1}{c}{$112 \times 112$} & 8                     & \begin{tabular}[c]{@{}c@{}}\specialrule{0em}{1pt}{1pt} $\text{3x3, 16-d LaConv} \times 2$  \end{tabular}      & \begin{tabular}[c]{@{}c@{}} \specialrule{0em}{1pt}{1pt} $\text{3x3, 16-d LaConv} \times 3$  \end{tabular}  \\          [4.5pt]
		\multicolumn{1}{c}{$56 \times 56$}   & -                     & 2x2, stride 2 pool                                                                                                               & 2x2, stride 2 pool                                                                                                       \\ [4.5pt]
		\multicolumn{1}{c}{$56 \times 56$}   & 4                     & \begin{tabular}[c]{@{}c@{}}\specialrule{0em}{1pt}{1pt} $ \text{7x7, 64-d LaConv}  \times 1$  \end{tabular}      & \begin{tabular}[c]{@{}c@{}} \specialrule{0em}{1pt}{1pt} $ \text{7x7, 64-d LaConv}  \times 3$  \end{tabular}  \\ [4.5pt]
		\multicolumn{1}{c}{$28 \times 28$}   & -                     & 2x2, stride 2 pool                                                                                                       & 2x2, stride 2 pool                                                                                                         \\ [4.5pt]
		\multicolumn{1}{c}{$28 \times 28$}   & 2                     & \begin{tabular}[c]{@{}c@{}}\specialrule{0em}{1pt}{1pt} $\text{7x7, 128-d LaConv} \times 2$ \end{tabular}              & \begin{tabular}[c]{@{}c@{}}\specialrule{0em}{1pt}{1pt} $\text{7x7, 128-d LaConv} \times 4$ \end{tabular} \\  [4.5pt]
		\multicolumn{1}{c}{$14 \times 14$}   & -                     & 2x2, stride 2 pool                                                                                                       & 2x2, stride 2 pool                                                                                                         \\ [4.5pt]
		\multicolumn{1}{c}{$14 \times 14$}   & 1                     & \begin{tabular}[c]{@{}c@{}}\specialrule{0em}{1pt}{1pt} $ \text{7x7, 256-d LaConv} \times 4$\end{tabular}        & \begin{tabular}[c]{@{}c@{}}\specialrule{0em}{1pt}{1pt} $ \text{7x7, 256-d LaConv} \times 6$\end{tabular} \\  [4.5pt]
		\multicolumn{1}{c}{$7 \times 7$}     & -                     & 2x2, stride 2 pool                                                                                                       & 2x2, stride 2 pool                                                                                                        \\[4.5pt]
		\multicolumn{1}{c}{$7 \times 7$}     & 1                     & \begin{tabular}[c]{@{}c@{}}\specialrule{0em}{1pt}{1pt} $ \text{7x7, 512-d }  \text{LaConv}  \times 1$ \end{tabular}   & \begin{tabular}[c]{@{}c@{}}\specialrule{0em}{1pt}{1pt} $ \text{7x7, 512-d }  \text{LaConv}  \times 3$ \end{tabular} \\  \midrule
		\multicolumn{4}{c}{Classifier}                                                                                                       \\ \botrule 
	\end{tabular} 
	\label{net}
	\vspace{-1em}
\end{table*}





\section{Experiments}
To validate LaConvNet and the LaConv module, we conduct extensive experiments on four benchmark datasets of Referring Expression Comprehension (REC) and Segmentation (RES) and Visual Question Answering (VQA), namely  RefCOCO~\cite{kazemzadeh2014referitgame}, RefCOCO+~\cite{kazemzadeh2014referitgame}, RefCOCOg~\cite{REFCOCOG}, Referit~\cite{kazemzadeh2014referitgame}, Flickr30k Entities~\cite{plummer2015flickr30k}, CLEVR~\cite{johnson2017clevr} and CLEVR-Ref+~\cite{liu2019clevr},  and compare them with a set of  latest methods~\cite{hudson2018compositional,shrestha2019answer,kim2018bilinear,Luo_2020_CVPR,yang2020improving}.
\subsection{Datasets and Metrics \label{datasets}}

\textbf{RefCOCO}~\cite{kazemzadeh2014referitgame},  \textbf{RefCOCO+}~\cite{kazemzadeh2014referitgame} and \textbf{RefCOCOg}~\cite{REFCOCOG} are three    datasets for  referring expression comprehension and segmentation, which are collected via an interactive game interface. RefCOCO and RefCOCO+ are splitted into \textit{train}, \textit{val}, \textit{test A} and \textit{test B}. The  referents of \textit{test A} are about people, while the ones of \emph{test B} are objects.  In contrast, RefCOCOg contains a \emph{val} and a \emph{test}.
In RefCOCO and RefCOCO+, there are 142\textit{k} expressions  for 50\textit{k} bounding boxes of 20\textit{k} images from MS-COCO~\cite{lin2014microsoft}. The expressions of RefCOCO  are mainly about absolute locations, while the ones of RefCOCO+ are more  about relative relations.   RefCOCOg has 104\textit{k} expressions for 54\textit{k} bounding boxes from 26\textit{k} images. The expressions of RefCOCOg are longer and more complex.

Following previous works~\cite{yang2019fast,Luo_2020_CVPR,LAVT}, IoU and IoU@0.5 are used to measure the performance of REC and RES tasks, respectively.  In particular, IoU determines the the Intersection-over-Union (\textbf{IoU}) between  the prediction and ground-truth.  For IoU@0.5,  when the IoU score is large than 0.5,  the prediction is regarded as correct. 

\textbf{Referit}~\cite{kazemzadeh2014referitgame} includes 20,000 images from SAIAPR-12 dataset~\cite{escalante2010segmented}. In Referit, there are 54,127, 5,872 and 60,103 image-expression pairs in  the training, validation   and testing sets, respectively. Compared to RefCOCO, RefCOCO+ and RefCOCOg, Referit contains more descriptions about the background.  We use IoU@0.5 as the metric for REC task.

\textbf{GRefCOCO}~\cite{GRES} is a large scale  dataset for generalized referring expression comprehension. It includes 90,022 multi-target expressions and 32,202 no-target expressions  for 19,994 images in the \textit{train}, \textit{val}, \textit{testA} and \textit{testB} splits. Since the \textit{testA} and \textit{testB} splits are not released, we report T-acc~\cite{GRES}, N-acc~\cite{GRES}, cIoU~\cite{GRES} and gIoU~\cite{GRES} on \textit{val} set.

\textbf{Flickr30k Entities}~\cite{plummer2015flickr30k} contains about 427,000 expressions for  31,7183 images,  29,783 images are used for training, 1000 for validation, and 1000 for testing. Compared to other REC datasets, the expressions of Flickr30k may refer to multiple objects. Following the common  settings~\cite{yang2019fast,transvg}, we merge all boxes of  referred objects  into a single one. IoU@0.5 is used as the metric.

\textbf{CLEVR}~\cite{johnson2017clevr} is a synthetic VQA dataset introduced by Johnson \textit{et al}~\cite{johnson2017clevr}, which aims to  examine  various reasoning skills, \emph{e.g.,} relation and  counting. It contains 999\textit{k} image-question pairs,  where 700\textit{k}, 150\textit{k} and 150\textit{k}  examples are  for training, validation and test, respectively. We use the classification accuracy as the metric.


\textbf{CLEVR-Ref+}~\cite{liu2019clevr} is a synthetic RES dataset derived from CLEVR.  It aims to  examine  referring expression comprehension with a set of reasoning tasks under the settings of one or multiple referents.   CLEVR-Ref+ contains 70K images and 700k expressions. We use the overall IoU to measure the accuracy.

\begin{table*}[t]
	\centering
	\caption{Comparisons of  LaConvNet and existing REC models on RefCOCO, RefCOCO+ and RefCOCOg. \#Params denotes the parameter size. * denotes the reproduced results.  ``cls-token'' and ``det-head'' denotes that the prediction head  of TransVG and LaConvNet, respectively.  }
	\setlength\tabcolsep{2.5pt}
	\begin{tabular*}{\textwidth}{@{\extracolsep{\fill}}lcccccccccc@{\extracolsep{\fill}}}
		\toprule
		\multicolumn{1}{c}{\multirow{2}{*}{Models}} & \multirow{2}{*}{\#Params} & \multirow{2}{*}{FLOPs} & \multicolumn{3}{c}{RefCOCO} & \multicolumn{3}{c}{RefCOCO+} & \multicolumn{2}{c}{RefCOCOg} \\
		\multicolumn{1}{c}{}                        &                         &                        & val      & testA   & testB   & val      & testA    & testB   & val           & test         \\ \midrule
		\textit{two-stage models:} \\
		CMN~\cite{hu2017modeling} & -  & - & - & 71.03 & 65.77 & - & 54.32 & 47.76 & - & - \\
		MAttNet~\cite{mattnet}                                      & 90M                     &  $\sim$81G                      & 76.65    & 81.14   & 69.99   & 65.33    & 71.62    & 56.02   & 66.58         & 67.27        \\
		NMTree~\cite{nmtree}  & -  & - & 76.41 & 81.21 & 70.09 & 66.46 & 72.02 & 57.52 & 65.87 & 66.44  \\
		CM-Att-Erase~\cite{cmatt}                                 &  -                       &       -                 & 78.35    & 83.14   & 71.32   & 68.09    & 73.65    & 58.03   & 67.99         & 68.67        \\ \midrule
				\textit{one-stage models:} \\
		RealGIN~\cite{realgin}                                      &     73M                    &            28.3G            & 77.25    & 78.70   & 72.10   & 62.78    & 67.17    & 54.21   & 62.75         & 62.33        \\
		FAOA~\cite{yang2019fast}                                        & 183M                    &   17.8G                     & 72.54    & 74.35   & 68.50   & 56.81    & 60.23    & 49.60   & 61.33         & 60.36        \\
		ReSC~\cite{yang2020improving}                                         & 182M                    &                       57.1G & 77.63    & 80.45   & 72.30   & 63.59    & 68.36    & 56.81   & 67.30         & 67.20        \\
		MCN~\cite{Luo_2020_CVPR}                                           & 73M                        &   44.5G                     & 80.08    & 82.29   & 74.98   & 67.16    & 72.86    & 57.31   & 66.46         & 66.01        \\
		LBYL-Net~\cite{lbyl} &115M&-&79.67 &82.91 &74.15& {68.64} &{73.38}& {59.49}  &-&-\\
		TransVG~\cite{transvg}                                      & 168M                    &     73.1G         & 81.02    & 82.72   & 78.35   & 64.82    & 70.70    & 56.94   & 68.67         & 67.73        \\ \midrule
				\textit{pre-trained models:} \\
		UNITER-L~\cite{uniter}                                     & 396M                    &  961.6G                      & 81.41    & 87.04   & 74.17   & 75.90    & 81.45    & 66.70   & 74.86         & 75.77        \\
		VILLA-L~\cite{villa}                                      & 399M                       &  1060.1G                      &  \underline{82.39}    & \underline{\textbf{87.48}}   & \underline{74.84 }  & \underline{\textbf{76.17}}    & \underline{\textbf{81.54}}    & \underline{\textbf{66.84 }}  & \underline{76.18 }        & \underline{76.71}        \\ 
		ViLT~\cite{kim2021vilt} (cls-token)* & 112M &42.0G& 79.51 &82.15&71.93&65.32&71.30&54.63&67.03&66.35\\
				ViLT~\cite{kim2021vilt} (det-head)* & 111M &42.0G&79.11 &81.90&73.21&68.16&73.35&57.35&67.16&66.47 \\ \midrule
		LaConvNet-S                                  & 24M                     & 8.9G                       & 80.26    & 83.51   & 75.82   & 69.82    & 73.61    & 60.98   & 74.45         & 74.54        \\
		LaConvNet-B                                  & 35M                     &  14.5G                      & \textbf{82.46}    & 84.66   & \textbf{78.16}   & 72.31    & 76.57    & 63.76   & \textbf{76.37}         & \textbf{76.75}        \\ \botrule
	\end{tabular*}
	\label{SOTA}
\end{table*}

\begin{table}[t]
	\centering
	\small
	\caption{Comparison of LaConvNet and existing REC models on Referit and Flickr. * denotes the reproduced results.}
	\setlength\tabcolsep{2pt}
	\begin{tabular*}{0.45\textwidth}{@{\extracolsep{\fill}}lcccc@{\extracolsep{\fill}}}
		\toprule
		\multicolumn{1}{c}{\multirow{2}{*}{Models}} & \multirow{2}{*}{\begin{tabular}[c]{@{}c@{}} \#Params\end{tabular}} & \multirow{2}{*}{\begin{tabular}[c]{@{}c@{}} FLOPs\end{tabular}}& Referit & Flickr \\
		\multicolumn{1}{c}{}                        &                                                                            && test    & test  \\ \midrule
		\textit{two-stage models:}\\
		MAttNet~\cite{mattnet}  &  90M                                 &    $\sim$81G                                                                  & 29.04   &  -     \\
		SimNet~\cite{simnet}   &      -                  &      -                                                                 & 34.54   & 60.89 \\
		DDPN~\cite{ddpn}        &      -                        &     -                                                                  & 63.00   & 73.30 \\ \midrule
		\textit{one-stage models:}\\
		ZSGNet~\cite{zsgnet}           &   -                    &     -                                                                   & 58.63   & 63.39 \\
		FAOA~\cite{yang2019fast}          &    183M                          &    17.8G                                                                    & 60.67   & 68.71 \\
		ReSC~\cite{yang2020improving}        &  182M                             &        57.1G                                                                & 64.60   & 69.28 \\
		TransVG~\cite{transvg}        &      168M                     &    73.1G                                                                   & \underline{70.73}   & \underline{79.10}\\ \midrule
		LaConvNet-S       &        24M                       & 8.9G                                                                      & {72.02}   & { 77.87} \\		LaConvNet-B        &   35M                          &        14.5G                                                               & \textbf{74.28}   & \textbf{79.95} \\ \botrule
	\end{tabular*}
 \vspace{-1em}
	\label{SOTA_fr}
\end{table}

\begin{table*}
\centering 
\caption{Comparison with the State-of-the-art methods on three RES datasets. The results colored in gray means that their visual backbone is  much large than LaConvNet.   * denotes the reproduced results. $\dagger$ denotes that large pre-trained language model is used as the text encoder. }
\setlength{\tabcolsep}{3pt} 
\begin{tabular}{cccccccccccc}
\toprule
\multirow{2}{*}{Models} & \multirow{2}{*}{\#Params} & \multirow{2}{*}{FLOPs} & \multirow{2}{*}{\begin{tabular}[c]{@{}l@{}}Visual \\ Backbone\end{tabular}} & \multicolumn{3}{c}{\centering RefCOCO} & \multicolumn{3}{c}{RefCOCO+} & \multicolumn{2}{c}{RefCOCOg} \\ \cline{5-12} 
& & && \multicolumn{1}{c}{val} & \multicolumn{1}{c}{testA} & testB & \multicolumn{1}{c}{val} & \multicolumn{1}{c}{testA} & testB & \multicolumn{1}{c}{val} & \multicolumn{1}{c}{test} \\ \midrule
\multicolumn{1}{c}{MAttNet~\cite{mattnet}} & 90M &$\sim$81G& MRCNN & \multicolumn{1}{c}{56.51} & \multicolumn{1}{c}{62.37} & \multicolumn{1}{c}{51.70} & \multicolumn{1}{c}{46.67} & \multicolumn{1}{c}{52.39} & \multicolumn{1}{c}{40.08} & \multicolumn{1}{c}{47.64} & 48.61 \\
\multicolumn{1}{c}{CMSA~\cite{CMSA}} & - & -& DR101 & \multicolumn{1}{c}{58.32} & \multicolumn{1}{c}{60.61} & \multicolumn{1}{c}{55.09} & \multicolumn{1}{c}{43.76} & \multicolumn{1}{c}{47.60} & \multicolumn{1}{c}{37.89} & \multicolumn{1}{c}{-} & - \\
\multicolumn{1}{c}{STEP~\cite{STEP}} & - & -& DR101 & \multicolumn{1}{c}{60.04} & \multicolumn{1}{c}{63.46} & \multicolumn{1}{c}{57.97} & \multicolumn{1}{c}{48.19} & \multicolumn{1}{c}{52.33} & \multicolumn{1}{c}{40.41} & \multicolumn{1}{c}{-} & - \\
\multicolumn{1}{c}{BRINet~\cite{BRINet}} & 241M & 367.6G& DR101 & \multicolumn{1}{c}{60.98} & \multicolumn{1}{c}{62.99} & \multicolumn{1}{c}{59.21} & \multicolumn{1}{c}{48.17} & \multicolumn{1}{c}{52.32} & \multicolumn{1}{c}{42.11} & \multicolumn{1}{c}{-} & - \\
\multicolumn{1}{c}{CMPC~\cite{CMPC}} & 118M & 126.6G& DR101 & \multicolumn{1}{c}{61.36} & \multicolumn{1}{c}{64.53} & \multicolumn{1}{c}{59.64} & \multicolumn{1}{c}{49.56} & \multicolumn{1}{c}{53.44} & \multicolumn{1}{c}{43.23} & \multicolumn{1}{c}{-} & - \\
\multicolumn{1}{c}{LSCM~\cite{LSCM}} & 128M & 130.4G& DR101 & \multicolumn{1}{c}{61.47} & \multicolumn{1}{c}{64.99} & \multicolumn{1}{c}{59.55} & \multicolumn{1}{c}{49.34} & \multicolumn{1}{c}{53.12} & \multicolumn{1}{c}{43.50} & \multicolumn{1}{c}{-} & - \\
\multicolumn{1}{c}{CMPC+~\cite{CMPC+}} & - & - & DR101 & \multicolumn{1}{c}{62.47} & \multicolumn{1}{c}{65.08} & \multicolumn{1}{c}{60.82} & \multicolumn{1}{c}{50.25} & \multicolumn{1}{c}{54.04} & \multicolumn{1}{c}{43.47} & \multicolumn{1}{c}{-} & - \\
\multicolumn{1}{c}{MCN~\cite{Luo_2020_CVPR}} & 73M & 44.5G& DN53 & \multicolumn{1}{c}{62.44} & \multicolumn{1}{c}{64.20} & \multicolumn{1}{c}{59.71} & \multicolumn{1}{c}{50.62} & \multicolumn{1}{c}{54.99} & \multicolumn{1}{c}{44.69} & \multicolumn{1}{c}{49.22} & 49.40 \\
\multicolumn{1}{c}{EFN~\cite{EFN}} & 232M & 124.7G& R101 & \multicolumn{1}{c}{62.76} & \multicolumn{1}{c}{65.69} & \multicolumn{1}{c}{59.67} & \multicolumn{1}{c}{51.50} & \multicolumn{1}{c}{55.24} & \multicolumn{1}{c}{43.01} & \multicolumn{1}{c}{-} & -\\
\multicolumn{1}{c}{BUSNet~\cite{BUSNet}} & - & -& DR101 & \multicolumn{1}{c}{63.27} & \multicolumn{1}{c}{66.41} & \multicolumn{1}{c}{61.39} & \multicolumn{1}{c}{51.76} & \multicolumn{1}{c}{56.87} & \multicolumn{1}{c}{44.13} & \multicolumn{1}{c}{-} & - \\
\multicolumn{1}{c}{CGAN~\cite{CGAN}} & 70M & 51.7G& DR101 & \multicolumn{1}{c}{64.86} & \multicolumn{1}{c}{68.04} & \multicolumn{1}{c}{62.07} & \multicolumn{1}{c}{51.03} & \multicolumn{1}{c}{55.51} & \multicolumn{1}{c}{44.06} & \multicolumn{1}{c}{51.01} & 51.69 \\
\multicolumn{1}{c}{LTS~\cite{LTS}} & - & -& DN53 & \multicolumn{1}{c}{65.43} & \multicolumn{1}{c}{67.76} & \multicolumn{1}{c}{63.08} & \multicolumn{1}{c}{54.21} & \multicolumn{1}{c}{58.32} & \multicolumn{1}{c}{48.02} & \multicolumn{1}{c}{54.40} & 54.25 \\
\multicolumn{1}{c}{VLT~\cite{VLT,vlt_cvpr}} & 89M & 142.5G& DN53 & \multicolumn{1}{c}{67.52} & \multicolumn{1}{c}{70.47} & \multicolumn{1}{c}{65.24} & \multicolumn{1}{c}{56.30} & \multicolumn{1}{c}{60.98} & \multicolumn{1}{c}{50.08} & \multicolumn{1}{c}{54.96} & 57.73  \\
{LAVT~\cite{LAVT}$\dagger$}*&{46M}&{87.8G}&{Swin-T}&\underline{69.27}&\underline{72.16}&\underline{65.59}&\underline{57.02}&\underline{62.52}&\underline{49.26}&\underline{58.76}&\underline{60.14}\\ 

\multicolumn{1}{c}{\textcolor{GREY}{ReSTR~\cite{ReSTR}}} & \textcolor{GREY}{122M} & \textcolor{GREY}{52.3G}& \textcolor{GREY}{ViT-B} & \multicolumn{1}{c}{\textcolor{GREY}{67.22}} & \multicolumn{1}{c}{\textcolor{GREY}{69.30}} & \multicolumn{1}{c}{\textcolor{GREY}{64.45}} & \multicolumn{1}{c}{\textcolor{GREY}{55.78}} & \multicolumn{1}{c}{\textcolor{GREY}{60.44}} & \multicolumn{1}{c}{\textcolor{GREY}{48.27}} & \multicolumn{1}{c}{\textcolor{GREY}{-}} &  \textcolor{GREY}{-} \\ 
\multicolumn{1}{c}{\textcolor{GREY}{LAVT~\cite{LAVT}$\dagger$}} & \textcolor{GREY}{118M}& \textcolor{GREY}{193.1G} & \textcolor{GREY}{Swin-B} & \multicolumn{1}{c}{\textcolor{GREY}{72.73}} & \multicolumn{1}{c}{\textcolor{GREY}{75.82}} & \multicolumn{1}{c}{\textcolor{GREY}{68.79}} & \multicolumn{1}{c}{\textcolor{GREY}{62.14}} & \multicolumn{1}{c}{\textcolor{GREY}{68.38}} & \multicolumn{1}{c}{\textcolor{GREY}{55.10}} & \multicolumn{1}{c}{\textcolor{GREY}{61.24}} & \textcolor{GREY}{62.09} \\
\multicolumn{1}{c}{\textcolor{GREY}{VLT~\cite{VLT}}$\dagger$} & \textcolor{GREY}{-}& \textcolor{GREY}{-} & \textcolor{GREY}{Swin-B} & \multicolumn{1}{c}{\textcolor{GREY}{72.96}} & \multicolumn{1}{c}{\textcolor{GREY}{75.96}} & \multicolumn{1}{c}{\textcolor{GREY}{69.60}} & \multicolumn{1}{c}{\textcolor{GREY}{63.53}} & \multicolumn{1}{c}{\textcolor{GREY}{68.43}} & \multicolumn{1}{c}{\textcolor{GREY}{56.92}} & \multicolumn{1}{c}{\textcolor{GREY}{63.49}} & \textcolor{GREY}{66.22} \\
\midrule
LaConvNet-S&24M &9.6G& None &69.74&72.21&67.29
&57.16&60.11&47.72&%
60.89&60.48
\\
LaConvNet-B &35M&15.7G& None&
\textbf{70.59}&\textbf{72.93}&\textbf{67.74}
&\textbf{60.43}&\textbf{64.47}&\textbf{52.02}&
\textbf{61.73}& \textbf{62.41}\\
\bottomrule
\end{tabular}
\label{res}
\end{table*}

\begin{table}[t]
	\centering
	\caption{Comparisons of  LaConvNet with existing approaches on Clevr-Ref+ dataset. \textit{Prog} denotes the number of extra program labels used  during  training. Since the FLOPs of IEP-Ref vary dynamically for different samples, we report the average FLOPs of 100 samples.  }
	\setlength\tabcolsep{6.0pt}
	\begin{tabular*}{0.45\textwidth}{@{\extracolsep{\fill}}lcccc@{\extracolsep{\fill}}}
		\toprule
		{Models}  & {\#Params} & FLOPs & Prog& {IoU}          \\ \hline
		RMI~\cite{rmi}     & -  &- & 0             & 56.1                 \\
		IEP-Ref~\cite{liu2019clevr}     & 49M & $\sim$11.2G   & 9K            & 76.0                 \\
		IEP-Ref~\cite{liu2019clevr}      &49M &$\sim$11.2G  & 18K           & 78.2                \\
		IEP-Ref~\cite{liu2019clevr}      & 49M &$\sim$11.2G & 700K          & 80.6                   \\
		IEP-Ref~\cite{liu2019clevr}      & 49M &$\sim$11.2G & GT            & \underline{ 81.6}     \\ \hline
		LaConvNet-S &23M&5.3G & 0             & {79.8}   \\
		LaConvNet-B &35M&8.6G & 0             & \textbf{82.3} \\ \bottomrule
	\end{tabular*}
	\label{clevrref} 
\end{table}

\begin{table}[t]
\centering 
\caption{Results of LaConvNet and existing methods on gRefCOCO \textit{val} set. Following the settings of RLA~\cite{GRES},  output masks with less than 50 positive pixels are regarded as non-target predictions. $\dagger$  denotes that no-target classifier is applied. }
	\setlength\tabcolsep{7.0pt}
\begin{tabular}{lcccc}
\toprule
Models      & cIoU  & gIoU  & N-acc & T-acc \\ \hline
MattNet~\cite{mattnet}     & 47.51 & 48.24 & 41.15 & 96.13 \\
LTS~\cite{LTS}         & 52.30 & 52.70 & -     & -     \\
VLT~\cite{VLT}         & 52.51 & 52.00 & 47.17 & 95.72 \\
CRIS~\cite{CRIS}        & 55.34 & 56.27 & -     & -     \\
LAVT~\cite{LAVT}        & 57.64 & 58.40 & 49.32 & 96.18 \\
RLA~\cite{GRES}         & -     & -     & 49.96 & 96.28 \\
RLA~\cite{GRES}$\dagger$        & \underline{62.42} & \underline{63.60} & \underline{56.37} & \underline{96.32} \\ \hline
LaConvNet-S & 57.06 & \textbf{62.83} & \textbf{59.16} & 99.65 \\
LaConvNet-B & \textbf{57.14} & 62.52 & 58.06 & \textbf{99.74} \\ \bottomrule
\end{tabular}
\label{gRefs}
\end{table}

\begin{table*}[t]
	\centering
	\caption{Comparisons of   LaConvNet and existing  methods on CLEVR. \#Params denotes the parameter size. * denotes the reproduced results.}
	\setlength\tabcolsep{5.0pt}
	\begin{tabular*}{\textwidth}{@{\extracolsep{\fill}}lcccccccc@{\extracolsep{\fill}}}
		\toprule
		\multirow{2}{*}{{Model}} & \multirow{2}{*}{{\#Params}} & \multirow{2}{*}{{FLOPs}}  &
		\multirow{1}{*}{{Overall}} & \multirow{2}{*}{{Count}} & {Cmp.} & \multirow{2}{*}{{Exist}} & {Query.} & {Cmp.}   \\
		&       &                        &   {Accuracy}                           &                                 & {Num.} &                                 & {Attr.}  & {Attr.}  \\ \midrule 
		BUTD~\cite{anderson2018bottom}*         &  66M & 8.9G  & 50.6 & 44.2& 68.7 & 64.3 &   44.5     & 53.7 \\
		Film~\cite{perez2018film}         &  -    & - & 97.6                              & 94.5                            & 93.8          & 99.2                            & 99.2            & {99.0}  \\
		RN~\cite{santoro2017simple}           & -   & -   & 95.5                              & 90.1                            & 93.6          & 97.8                            & 97.1            & 97.1      \\
		BAN~\cite{kim2018bilinear}*           & 147M  & 19.1G &92.2&88.3&94.9&96.4&91.2&94.7  \\
		RAMEN~\cite{shrestha2019answer}*            &  76M & 8.1G &  87.8&82.1&83.3&93.9&90.6&87.0 \\
		DDprog~\cite{suarez2018ddrprog}&-&-&98.3&96.5&98.4&98.8&99.1&99.0\\
		Tbdnet~\cite{mascharka2018transparency}&-&-&98.7&96.8&99.1&98.9&99.4&99.2\\
		NS-CL~\cite{mao2019neuro}&-&-&98.9&\underline{\textbf{98.2}}&99.0&98.8&99.3&99.1\\
		MACNet~\cite{hudson2018compositional}            & 39M &8.0G& \underline{98.9}& {97.1} & \underline{99.1 }&\underline{99.5}&\underline{99.5}&\underline{\textbf{99.5}}  \\ 
		\midrule
		LaConvNet-S &   19M  & 1.9G & 98.3   & 96.4         &   97.5   & 99.2       & 99.3       & 98.3  \\ 
		LaConvNet-B &     31M & 3.7G &\textbf{ 99.1}  & { 97.9 }    & \textbf{99.4  }     & \textbf{99.5 }      &  \textbf{99.6 }     &  {99.3}\\ \botrule
	\end{tabular*}
	\label{clevr_s}
\end{table*}

\begin{table}[t]
\caption{Comparison of LaConvNets and other multi-task VL models on VQAv2 and SNLI-VE. UniT is a VL transformer  that can  simultaneously learn different tasks. We report accuracy on \textit{test-dev} set of VQAv2  and  \textit{test} set of SNLI-VE. }
\setlength{\tabcolsep}{5pt} 
\begin{tabular}{lccc}
\toprule
\multicolumn{1}{c}{Methods} & \#Params   & VQAv2 & SNLI-VE \\ \midrule
Unified-IO$_\texttt{base}$~\cite{unified-io} &241M& 61.8&-\\
UniT-single-task~\cite{Unit}            & 201M      & 66.38 & 70.52   \\
UniT-shared~\cite{Unit}                   & 201M      & 66.97 & 73.16   \\ \midrule
LaConvNet-S                 & $\sim$25M & 66.58 & 71.77   \\
LaConvNet-B                 & $\sim$36M & \textbf{67.21} & \textbf{}74.21   \\ \bottomrule
\end{tabular}
\label{vqa}
\end{table}

\begin{table*}[t]
	\centering
			\caption{Performance comparison between the modular network  with LaConv  and other methods  on REC   and VQA. \textit{Prog.} denotes the number of extra program ground truths used during   training. \textit{LaConv}$\times$6 is a structure with 6 LaConv layers.  CDN and R101 denote the visual backbones, \emph{i.e.,} CspDarkNet and ResNet101, respectively.}
	\setlength\tabcolsep{7.0pt}
	\begin{tabular*}{\textwidth}{ccccc|ccc}
		\toprule
		\multicolumn{5}{c|}{REC Task}                                                                                                         & \multicolumn{3}{c}{VQA Task}                                                                               \\ \midrule
		\multicolumn{1}{l}{\multirow{2}{*}{Models}} & \multicolumn{2}{c}{RefCOCO}                         & \multicolumn{2}{c|}{RefCOCO+}   & \multicolumn{1}{l}{\multirow{2}{*}{Models}} & \multicolumn{1}{c}{\multirow{2}{*}{Prog.}} & CLEVR       \\
		\multicolumn{1}{c}{}                        & testA          & \multicolumn{1}{c}{testB}          & testA          & testB          & \multicolumn{1}{c}{}                        & \multicolumn{1}{c}{}                       & Accuracy      \\ \midrule
		\multicolumn{1}{l}{NMTree~\cite{nmtree}}          & 81.21          & \multicolumn{1}{c}{70.09}          & 72.02          & 57.52          & \multicolumn{1}{l}{DDprog~\cite{suarez2018ddrprog}}          & \multicolumn{1}{c}{700k}                   & 98.3          \\
		\multicolumn{1}{l}{CM-Att-Erase~\cite{cmatt}}    & \underline{83.14}    & \multicolumn{1}{c}{71.32}          & 73.65          & 58.03          & \multicolumn{1}{l}{Tbdnet~\cite{mascharka2018transparency}}          & \multicolumn{1}{c}{700k}                   & 98.7          \\
		\multicolumn{1}{l}{MCN~\cite{Luo_2020_CVPR}}             & 82.29          & \multicolumn{1}{c}{74.98}          & 72.86          & 57.31          & \multicolumn{1}{l}{MACNet~\cite{hudson2018compositional}}          & \multicolumn{1}{c}{0}                      & {\underline{98.9}}    \\
		\multicolumn{1}{l}{TransVG~\cite{transvg}}         & 82.72          & \multicolumn{1}{c}{\underline{78.35}}    &\underline{70.70}   & \underline{56.94}    & \multicolumn{1}{l}{NS-CL~\cite{mao2019neuro}}           & \multicolumn{1}{c}{0}                      & {\underline{98.9}}    \\ \midrule
		\multicolumn{1}{l}{LaConv$\times$6+CDN}     & \textbf{86.02} & \multicolumn{1}{c}{\textbf{79.49}} & \textbf{77.24} & \textbf{62.53} & \multicolumn{1}{l}{LaConv$\times$6+R101}    & \multicolumn{1}{c}{0}                      & \textbf{99.1} \\ \botrule
	\end{tabular*}
\label{modular}
\vspace{-1em}
\end{table*}

\begin{table}[t]
\caption{Ablation study of dynamic filters and parameter generation in LaConv. We conduct all experiments based on LaConvNet-S. \textit{LPG} denotes the language-conditioned parameter generations.  }
	\setlength\tabcolsep{4.0pt}
\begin{tabular*}{0.45\textwidth}{@{\extracolsep{\fill}}lccc@{\extracolsep{\fill}}}
\toprule
Settings                        & Params & FLOPs & RefCOCO \\ \midrule
\textit{dynamic filters:}       &        &       &          \\
  group=1                     &   22M     &  8.0G     &   77.02       \\
  kernel=3                    &   22M     &  8.1G     &   77.78       \\
    kernels=[3,7,7,7,7]                    &   24M     &   8.9G    &  80.26           \\ \midrule
\textit{parameter generation:} &        &       &          \\
no pixel packing                &   22M     &   9.1G    &   79.12       \\
language only                   &   19M     &    6.4G   &     75.58     \\ 
LPG + pixel packing  &   24M     &   8.9G    &  80.26         \\ \botrule
\end{tabular*}
\label{ablation}
\end{table}

\subsection{Experimental Settings \label{detail}}
\noindent \textbf{LaConvNet.} We use Glove~\cite{pennington2014glove} word embeddings with a dimension of 300 to represent each input word. The language encoder is built with an LSTM network and  a self-attention layer~\cite{vaswani2017attention},  and their dimension are set to  512. 
The input images are resized to $416 \times 416$, $320\times 320$ and $224\times 224$ for REC, RES and VQA, respectively.
 The task-specific heads of LaConvNet are directly borrowed from previous works~\cite{yu2019deep,Luo_2020_CVPR}.  For REC, the output three-scale  visual features are firstly fused by a simple multi-scale fusion module~\cite{Luo_2020_CVPR}, upon which a detection head~\cite{yolov3} is used to predict the target box. For VQA, we  first attentively pool the last-layer visual features and the language features and then fuse them by additions, followed by an MLP to predict the answer.

All models are trained by  \emph{Adam} optimizer~\cite{kingma2014adam} with an initial learning rate of 1e-4. 
For REC task, the total training epochs are set to 25, 3 of which are for warm-up, and the learning rate is decayed by cosine schedule.  We pre-train LaConvNets on VG+COCO+Flickr (200k images) with 10 epochs. 
For CLEVR and CLEVR-ref+, the number of training epochs is 23, and 3 epochs are for warm-up, and the learning rate is decayed by a factor of 0.2 at the 20-\textit{th} and the 22-\textit{th} epochs. For CLEVR and CLEVR-ref+, LaConvNets are trained from scratch. 

\noindent \textbf{ViLT~\cite{kim2021vilt}.} To adopt ViLT to  REC task, we try two  types of  REC heads~\cite{transvg,Luo_2020_CVPR} on ViLT.  In the first setting, we uses a 4-layer MLP to predict the box based on the [CLS] token~\cite{transvg}. In the other setting, ViLT predicts the box through an anchor-based detection head~\cite{yolov3}.   Fine-tuning ViLT takes 60 epochs with a learning rate of 1e-4.  We use AdamW as the optimizer. Following \cite{transvg}, we  apply some data augmentations to avoid  overfitting.

\subsection{Experimental Results}

\subsubsection{Quantitative Analysis}
\label{discuss_rec}
\noindent \textbf{Referring Expression Comprehension (REC).}
In Tab.~\ref{SOTA} - \ref{SOTA_fr}, we compare LaConvNets with existing REC models on five common REC benchmark datasets \emph{i.e.,}, RefCOCO, RefCOCO+, RefCOCOg, ReferIt and Flick30K.   The first observation  from these tables  is that LaConvNet-S can already outperform existing one-stage and two-stage approaches by  notable margins,  \emph{e.g.},  +5.78\% over TransVG~\cite{transvg} and +6.46\% over CM-Att-Erase~\cite{cmatt}.   To explain,  visual features of these modular networks are usually redundant and noisy~\cite{mae}, and inevitably hurt the efficiency of multi-modal reasoning. By contrast, LaConvNets directly extract language-relevant visual features from raw pixels, greatly improving the quality and compactness of visual features. 
As the model size increases, LaConvNet-B  can obtain obvious performance gains, suggesting its superior scalability.    Notably,   the performance gains  of LaConvNets are more obvious  on more challenging dataset, \emph{i.e.,} RefCOCOg,  confirming  its superior   ability  in visual reasoning.    On ReferIt and Flickr, whose data distributions are  greatly different from  the  RefCOCO-series, LaConvNets can also outperform existing  REC models  with significant improvements.   In addition to  performance, we can also see the   superiorities of LaConvNets in   parameter  size and model efficiency.  In particular, the total parameters of LaConvNet-S are 1/7 of   TransVG. In terms of FLOPs, the total cost of LaConvNet-S is even 262 times  cheaper than the one of TransVG.  Compared to other one-stage REC models, the same advantages can be still witnessed,  confirming the great potential  of LaConvNets on edge devices.

In Tab.~\ref{SOTA}, we also compare LaConvNets with large-scale pre-trained models, \emph{i.e.,} UNITER~\cite{uniter}, VILLA~\cite{villa} and ViLT~\cite{kim2021vilt}. All these  large-scale models  require  about 4M images for pre-training.  By contrast, LaConvNets use much fewer  pre-training data, \emph{i.e.,}  200k images, but their performance is still comparable with  those large-scale pre-trained models, \emph{e.g.,} +0.19\% gains over VILLA-L on RefCOCOg.   Compared to the existing  unified model  ViLT, which is also pre-trained on large-scale data,   LaConvNets not only  performs better than ViLT, but also have much  cheaper   computational costs  than ViLT, \emph{i.e.,} 8.9G FLOPs \textit{vs.} 42.0G FLOPs.   

Overall, these results  
reveal that  LaConvNet, as a multi-modal and end-to-end network, is   capable of  accurate vision-and-language alignments. It also confirms that LaConvNet can be  a viable alternative to existing modular networks.

\noindent \textbf{Referring Expression Segmentation (RES).}
In Tab.~\ref{res} and \ref{clevr_s}, we compare LaConvNets with existing RES models on RefCOCO, RefCOCO+,  RefCOCOg and CLEVR-ref+.   In Tab.~\ref{res} , our experiments show that existing RES models incur high computational costs. For instance, VLT achieves state-of-the-art performance but  requires a  computational overhead of 142.5G FLOPs.  LAVT~\cite{LAVT}   directly embeds language information into visual backbone, but its computational cost remains expensive due to its large visual backbone  and heavy multi-modal fusion blocks.  In particular, when LAVT uses Swin-B~\cite{swintransformer} as the visual backbone,  its FLOPs can be up to 193.1G FLOPs. Even using the smallest size of Swin Transformer as the visual backbone, \emph{i.e.,} Swin-T~\cite{swintransformer},  the parameters and computation costs of  LAVT also exceed those of LaConvNet-S by 2 times and 9 times, respectively.  
Compared to these methods, LaConvNets achieve better trade-offs between efficiency and performance under the similar experimental settings, \emph{e.g.,} the text encoder and the visual encoder with comparable parameter size.     For instance, LaConvNet-S achieves better results than VLT (DN53)  while requiring only 6.7\% FLOPs and 27\% parameters of VLT.  As the model size increases, LaConvNet-B outperforms LAVT (Swin-T) by 2.15\%, 3.41\%, and 2.97\% on RefCOCO, RefCOCO+, and RefCOCOg, respectively.  Although LAVT (Swin-B) and VLT (Swin-B) can outperform LaConvNets with a  large visual backbone and text encoder, their computational costs can be up to 20 times that of LaConvNet-S. Overall, we believe that LaConvNets can still be an efficient alternative to these large models for RES tasks.

In Tab.~\ref{clevrref}, we compare LaConvNets with other approaches on the synthetic RES dataset called Clevr-Ref+. This dataset contains longer and more complex expressions, which require higher reasoning ability. We observe that LaConvNets outperform other compared approaches with much fewer parameters and computations. Compared to RMI, the end-to-end RES model, the performance gains of LaConvNets are significant, \emph{i.e.,} up to +26.2\% IoU. Although the symbol-based approaches, \emph{e.g.,} IEP-Ref, use all ground-truth programs, their performances are still inferior to LaConvNet-B. These results further confirm the effectiveness and efficiency of LaConvNets.

In Tab.~\ref{gRefs}, we validate the robustness of LaConvNets on gRefCOCO~\cite{GRES}, which contains single-object, multi-object and no-target expressions. From the results, we firstly observe that LaCovNet-S achieves 59.16\% N-acc and 99.65\% T-acc on gRefCOCO val set, which outperforms the state-of-the-art model, \emph{i.e.,} RLA~\cite{GRES}, by +2.79\% and +3.33\%, respectively. Such results suggest that LaConvNets can well handle no-target expressions.  In terms of  multi-object expressions, LaConvNets also have competitive performance. For instance, LaConvNet-S achieves 62.83\% gIoU on gRefCOCO \textit{val} set, which suppresses most existing RES models~\cite{VLT,CRIS,LAVT}. Overall, these results greatly confirm the   robustness of LaConvNets.



\noindent \textbf{Visual Question Answering (VQA).}
To validate the generalization ability of LaConvNets, we also apply LaConvNets to VQA task, of which results are given in Tab.~\ref{clevr_s} and \ref{vqa}.  In Tab.~\ref{clevr_s}, we compare LaConvNets with existing methods on CLEVR. From it we can see that LaConvNet-B achieves superior overall accuracy than existing approaches and also demonstrates better parameter efficiency. On multiple metrics, LaConvNet-B  reaches the performance upper bound of the dataset, showing its strong reasoning ability. 

We further compare  LaConvNets with the multi-task Transformer model, \emph{i.e.,} UniT~\cite{Unit}, on two real-world VQA benchmark datasets. {In  particular, UniT is a multi-modal and multi-task network, which can simultaneously conduct tasks of different modalities.  In terms of network architecture, they are still modular and require independent visual backbones and another fusion branches. In contrast, LaConvNets jointly models   vision-language information in one unified multi-modal network from raw pixel level, greatly reducing computational overhead.} On VQAv2 and SNLI-VE, UniT requires 201M parameters, about 5 times that of LaConvNet-B. In term of performance, LaConvNets also achieve better results, \emph{i.e.,} +0.24 on VQAv2 and +1.05 on SNLI-VE. {These results  demonstrate the high efficiency and generalization ability of LaConvNets. We believe that LaConvNets can be a good supplement to existing VL research which mainly pursue giant Transformer-based models to break the limit of VL learning ability. }

\subsubsection{Ablation Study}
To gain deep insights of LaConvNets, we provide the detailed ablation results in Tab.~\ref{modular}-\ref{multimodal}.

\begin{table}[t]
	\centering
	\caption{Comparisons of  LaConv block and other multi-modal modules. To make a fair comparison, we directly replace all LaConv blocks of  LaConvNet-S with these multi-modal modules.   }
	\setlength\tabcolsep{3.5pt}
	\begin{tabular*}{0.45\textwidth}{lcccc}
		\toprule
		\multicolumn{1}{l}{\multirow{2}{*}{Modules}} & \multirow{2}{*}{Params} & \multirow{2}{*}{FLOPs} & \multirow{2}{*}{RefCOCO} & \multirow{2}{*}{RefCOCO+} \\
		\multicolumn{1}{c}{}                  &                         &                        &                          &                           \\ \midrule
		LaConv                                 &  23.1M                       &   8.9G                     &   80.26                       &       69.82                    \\ \midrule
		FiLM~\cite{perez2018film}                                   &   30.4M                      &   9.8G                     &    78.95                    &   68.78                        \\
		QGHC~\cite{gao2018question}                                   &  55.7M                       &   6.8G                     &      76.26                    &  66.11                         \\
  \botrule
	\end{tabular*}
	\label{multimodal}
\end{table}

\noindent \textbf{Ablations of LaConv.} In Tab.~\ref{ablation}, we provide four different settings for the dynamic filters and parameter generation of LaConv:
\begin{itemize} 
	\item ``Group=1'' denotes that all convolution groups are set to 1 in   LaConvNet.
	\item ``Kernel=3'' denotes kernel size of LaConv. 
	\item ``No pixel packing'' denotes that we remove  pixel packing in the language-conditioned parameter generation, as mentioned in Sec~\ref{pixelpacking}.
	\item ``Language only'' denotes that the convolution kernels are directly predicted by the language features. In this case, the generated  convolution kernels are shared for all regions.
\end{itemize}
From Tab.~\ref{ablation}, we observe that fewer convolution groups and smaller convolution kernels will  hinder model performance. To explain,   more convolution groups can often improve the model capacity and larger convolution kernels can increase the receptive fields. Such observations are also consistent with those of  existing conventional convolution networks~\cite{tan2019efficientnet}. In Tab.~\ref{ablation}, we also validate  different settings of parameter generation.  One can be seen that when pixel packing is removed, the model performance declines and the computational costs increase. Such results also  validate the problem of \emph{low-rank degeneration}~\cite{bhojanapalli2020low} in parameter generations.    From the results of ``\emph{language only}'', we can see the effectiveness of the condition matrix design. Without the transformation of condition matrix,  the generated convolution kernels are image-inrelevant and shared for all visual regions. As it can be seen, this alternative design makes the model decline greatly in performance. It also  suggests the importance of the proposed language-conditioned parameter generation.

\begin{table}[t]
\centering
\caption{Parameters and FLOPs of LaConv  and other modules. ``VP'' and ``MF'' denotes the visual processing module in visual backbone and multi-modal module in fusion branch, respectively. We calculate the costs of a single layer. For VLT,  ``MF'' only calculates the costs of Transformer decoder.  }
	\setlength\tabcolsep{3pt}
\begin{tabular}{l|ccc|ccc}
\toprule
\multicolumn{1}{c|}{\multirow{2}{*}{Models}} & \multicolumn{3}{c|}{\#Param}               & \multicolumn{3}{c}{FLOPs}                 \\  
\multicolumn{1}{c|}{}                        & VP    & \multicolumn{1}{c|}{MF}   & Total & VP   & \multicolumn{1}{c|}{MF}    & Total \\ \midrule
VLT~\cite{VLT}                                         & 5.2M  & \multicolumn{1}{c|}{4.1M} & 9.3M  & 1.0G & \multicolumn{1}{c|}{2.96G} & 3.96G \\
LAVT~\cite{LAVT}                                         & 12.5M & \multicolumn{1}{c|}{7.8M} & 20.3M & 3.8G & \multicolumn{1}{c|}{1.0G}  & 4.8G  \\
ViLT~\cite{kim2021vilt}                                         & \multicolumn{2}{c|}{7.0M}         & 7.0M  & \multicolumn{2}{c|}{2.1G}         & 2.1G  \\ \midrule
LaConvNet                                   & \multicolumn{2}{c|}{1.4M}         & 1.4M  & \multicolumn{2}{c|}{0.45G}        & 0.45G \\ \bottomrule
\end{tabular}
\label{flops_module}
\end{table}

\noindent \textbf{Comparison of different multi-modal convolutions.}  In Tab.~\ref{multimodal}, we compare LaConv with  other multi-modal convolutions in the architecture of LaConvNet,   \emph{i.e.}, FiLM~\cite{perez2018film} and QGHC~\cite{vaswani2017attention}.   In particular,  FiLM directly fuses language information into the visual processing via language-conditioned normalization layers.  This principle  also closes to that of  LAVT~\cite{LAVT}. 
 These approaches still rely on additional heavy multi-modal fusion blocks, which limit their efficiency.  Compared to them, LaConv unify the multi-modal fusion and visual processing into one lightweight block, leading to better performance and efficiency, as shown in  Tab.~\ref{multimodal}.   QGHC~\cite{gao2018question}, which also predicts convolution kernels by language features,  takes fewer FLOPs than LaConv, but its performance is still far behind LaConv. In summary, compared to these multi-modal modules, LaConv can achieve better trade-offs between efficiency and effectiveness.

\noindent \textbf{Results of LaConv in modular networks.}  In Tab.~\ref{modular}, we also validate LaConv as a plug-in module to the existing modular  network, of which results are given in Tab.~\ref{modular}. Specifically, we construct a modular network with 6 LaConv layers and a visual backbone.   From this table, we can see that as a multi-modal inference module, LaConv still outperforms existing methods on REC and VQA.  For example, LaConv$\times$6+CDN outperforms  TransVG by +7.17\% on RefCOCO+.  On VQA,  the same advantages can be also witnessed.  Compared to the symbol-based approach, \emph{i.e.,} Tbdnet~\cite{mascharka2018transparency}, which uses all program ground-truth during training, LaConv still outperforms this method by +0.2\% without additional program annotations. Notably, this performance  almost reaches the upper-bound of CLEVR. These results demonstrate the strong generalization ability of LaConv as an plug-and-play module for VL tasks.

\begin{figure}[t]
	\centering
	\includegraphics[width=1\columnwidth]{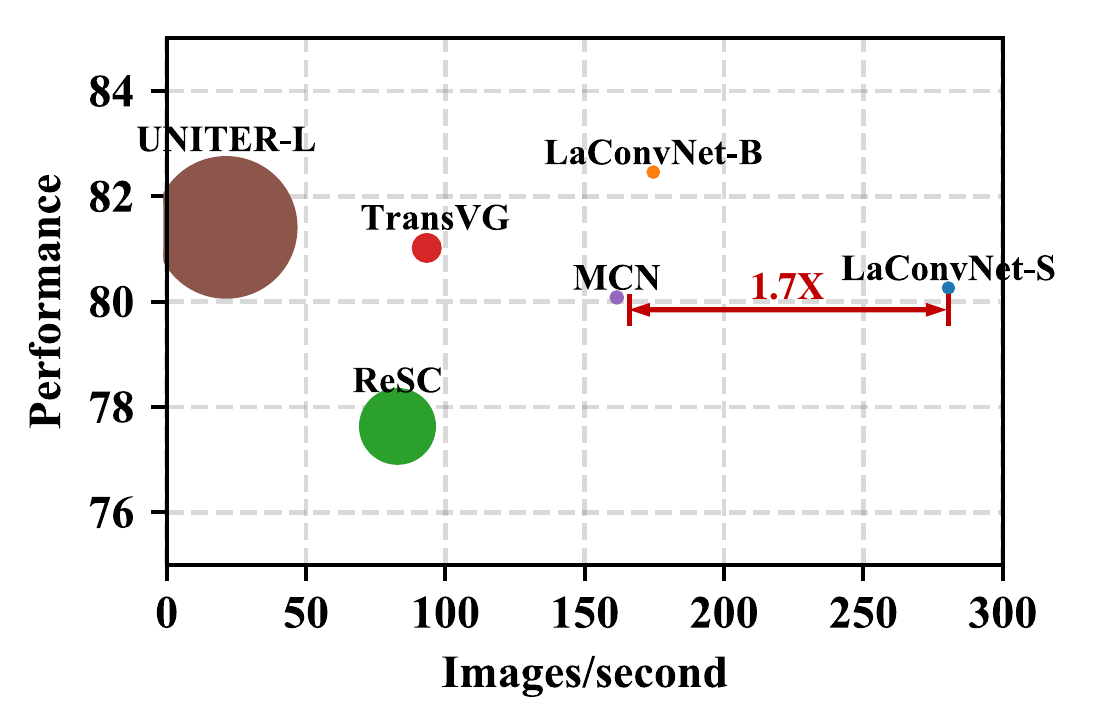}
	
	\caption{The   GPU latencies and memory of LaConvNets and other models on RefCOCO. The point size  corresponds to the GPU memory  footprint.  All models are tested on a NVIDIA A100 GPU with a mini-batch of 64. }
	\label{vis_speed} 
\end{figure}

\begin{figure*}[t]
	\centering
	\includegraphics[width=2\columnwidth]{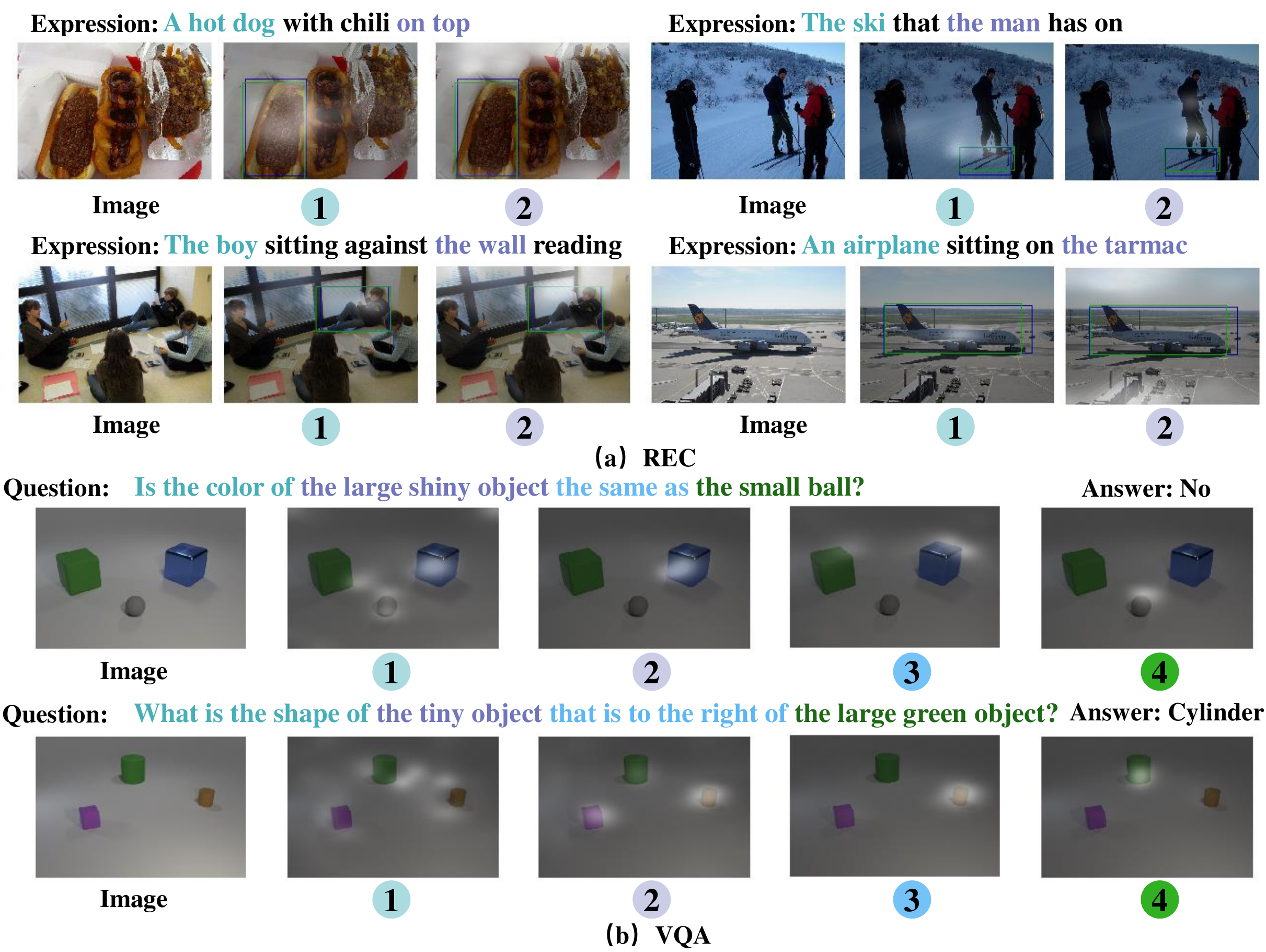}
	
	\caption{Visualizations of the attention maps in language-conditioned parameter generation.   We visualize  phrases of a question and their corresponding visual attention. In language-conditioned parameter generation, the phrase and its corresponding visual regions are accurately aligned. }
	\label{vis1}
	\vspace{-1em}
\end{figure*}

\subsubsection{Model Efficiency}
The  efficiency of {LaConvNet} is  depicted  in Tab.~\ref{SOTA}-\ref{vqa} and \ref{flops_module},  and Fig.~\ref{vis_speed}. In Tab.~\ref{SOTA}-\ref{vqa}, we compare LaConvNets  with existing models in terms of   parameter size and  computation, \emph{i.e.}, FLOPs.  As a language-conditioned convolution network, {LaConvNets} are much more lightweight and efficient than existing networks, as discussed in Sec.~\ref{discuss_rec}.     

In Tab. \ref{flops_module}, we compare the costs of LaConv and other modules in existing methods~\cite{VLT,LAVT,kim2021vilt}. By comparison, we find that existing multimodal fusion methods require much more expensive parameters and computation than LaConvNet. For example, the fusion block of VLT requires 2.96G FLOPs, about 6.5 times that of LaConvNet.  Meanwhile, most existing approaches still require additional costs for the visual processing module,  \emph{e.g.,} 2.96G FLOPs of VLT, which further reduce their efficiency. 

In Fig.~\ref{vis_speed},   we further present the  practical GPU throughputs and   memory of  LaConvNets.  Compared to large-scale pre-trained model, \emph{i.e.,} UNITER~\cite{uniter}, LaConvNets are superior in both throughputs and memory costs.  We also see that LaConvNet-S is  more efficient than existing one-stage models, \emph{e.g.,} 1.7 times faster than MCN~\cite{Luo_2020_CVPR},  which is a real-time one-stage REC model.   These experiments well support the efficiency of LaConvNets.


%


\subsubsection{Qualitative Analysis}
In this section, we give  detailed visualizations to answer two key questions of LaConv and LaConvNet, \emph{i.e.,} ``\textit{is the parameter generation reliable and interpretable?}''  and ``\textit{what convolutions are learned from the natural language instructions?} '' Besides, we also provide   typical failure cases  to  further analyze LaConvNets.

\begin{figure*}[t]
	\centering
	\includegraphics[width=2\columnwidth]{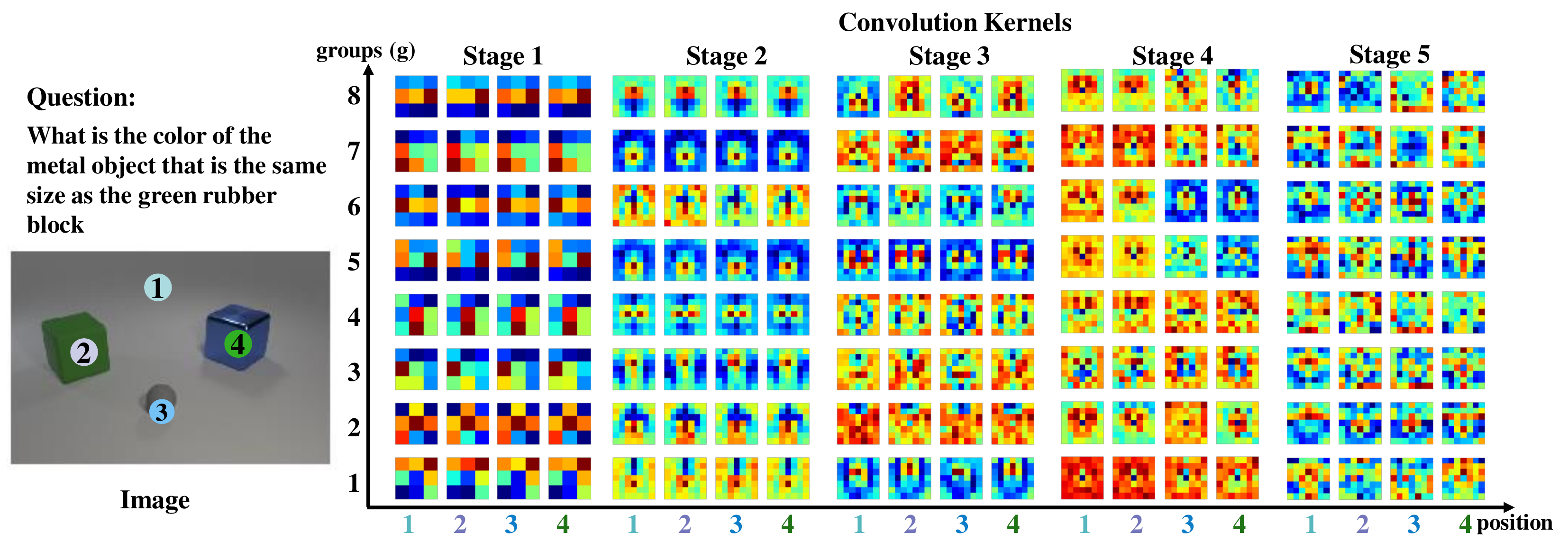}
	
	\caption{Visualizations of  the dependences of text phrases and their corresponding image regions during language-conditioned parameter generation.
		The colors denote the magnitude of values in filters, and  the  redder  color  indicates a larger value. 
		For each stage, we select the last layer for visualization. }
	\label{vis2}
	\vspace{-3mm}
\end{figure*}

\begin{figure*}[t]
	\centering
	\includegraphics[width=2\columnwidth]{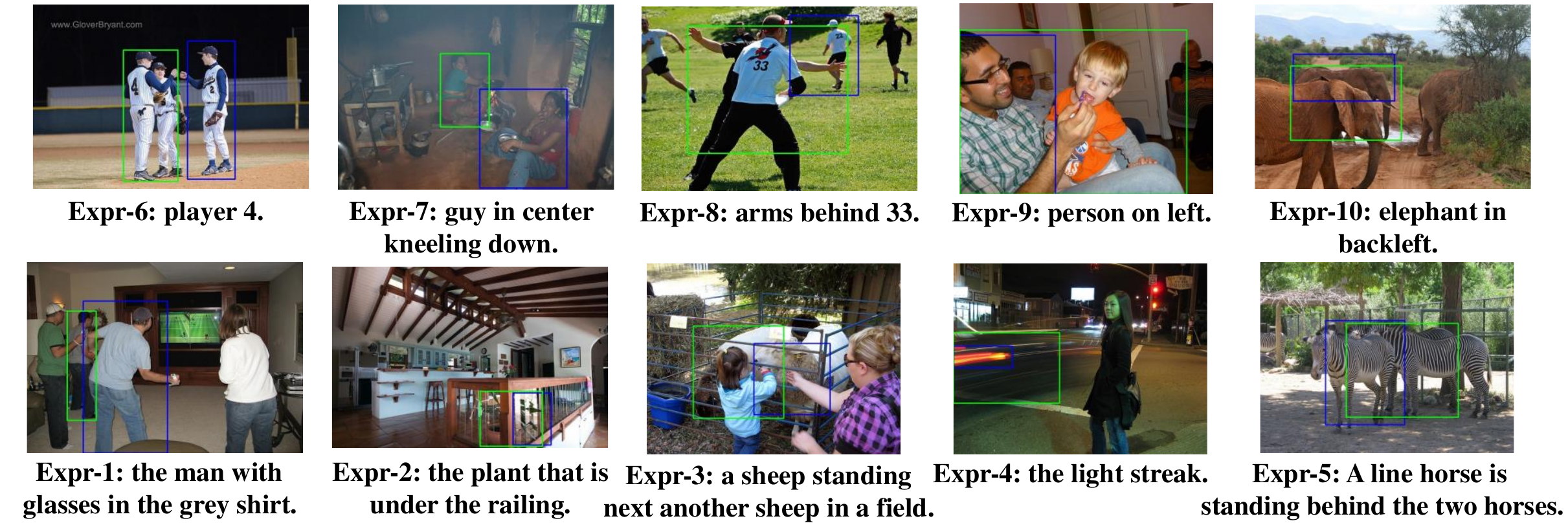}
	
	\caption{ Failure cases of    LaConvNet-S on RefCOCO+ (top row) and RefCOCOg (bottom row). The blue   and green boxes are the ground-truths and predictions, respectively. }
	\label{vis3}
	\vspace{-1em}
\end{figure*}

\textbf{Is the condition generator reliable and interpretable?}
{LaConv} is more interpretable than the traditional convolution due to the language-dependent parameter generation.
To support this argument, we we select the last convolution layer of LaConvNets to visualize the affinity matrix $A$ in the parameter generations.  
As shown in Fig~\ref{vis1}, each phrase of a text attends to the corresponding region.
For instance, in the first example of Fig.\ref{vis1} (a),  the different phrases of ``a hot dog'' and ``on top'' highlights the corresponding regions.
Analogically, the spatial phrase of ``the right of'' in the second example of of Fig.\ref{vis1} (b) is also related to the \textit{right} object. 
In addition, other referring phrases, \emph{e.g.,} ``the large shiny object'' and ``the tiny object'', are also visualized in the attention maps.
Based on these observations, we  believe that the generated convolution filters of a position can accurately execute the corresponding language instructions, which makes {LaConv} more reliable and  interpretable.

\textbf{What convolutions are learned from language instructions?}
Unlike the conventional convolutions that are weight-sharing for each spatial position, {LaConv}  depends on both image position and text content.
To examine its dynamics, we visualize the filters in each stage of {LaConvNet} in Fig.\ref{vis2}.
For a better understanding, we select four positions of the example image, and visualize their filters.
From Fig.\ref{vis2}, the first observation is that the filters for different groups vary greatly (from the vertical axis), which suggests that each group of convolutions is responsible for different recognition patterns.
The second observation is that the filters at the initial stage are relatively static, \emph{i.e.,} filters of the same group are similar for different positions (read from the horizontal axis).
Such a finding suggests that these convolutions focus on learning low-level visual representations, and they  are less affected by natural language information.
However, we notice that as the recognition progresses, the filters of the same group  vary drastically, presenting different intensities and morphs, \emph{e.g.,} Stage3-5.
This observations suggest that the language instructions start to dynamically guide the visual recognition,  so different positions present distinct patterns.  Conclusively, these visualizations indicate that the language-guided visual recognition of LaConvNet is a continuous process, and the impact of  language information can be reflected on its weights. 

\textbf{Failure cases.} In Fig.~\ref{vis3}, we give  the  typical failure cases on REC.  We observe that   some failure examples occur due to the ambiguous annotations, \emph{e.g.,} Expr-2, Expr-6 and Expr-9. Meanwhile, visual occlusion is also a  main factor for  incorrect recognition, \emph{e.g.,} Expr-5, Expr-8 and Expr-10. In addition, we also find that LaConvNets  fail to accomplish very abstract  descriptions, \emph{e.g.,} ``player 4'' and `` arms behind 33''.  However, these types of examples are not common in existing datasets. And they has gone beyond the scope of the conventional REC research.

\section{Conclusion}
In this paper, we establish the first fully language-driven convolution network for vision-and-language tasks, termed LaConvNet, which can get rid of large visual backbones and achieve language-guided visual recognition.  Specifically, LaConvNet is built by a novel dynamic convolution module called LaConv. The convolution kernels of LaConv are predicted from text features, so it can achieve differentiated visual feature learning for different natural language instructions. This property also enable LaConv to perform visual modeling and multi-modal inference in one processing step. 
Based on {LaConv}, we can build the unified and end-to-end LaConvNet. 
{LaConvNet} gets rid of CNN blocks entirely  and directly performs visual reasoning  from raw pixels, which greatly differs from existing modular networks for VL tasks.  Extensive experiments are conducted on seven benchmarks of three VL tasks.  The  experimental results  not only show the comparable or even better performance of LaConvNet against existing modular multi-modal networks, but also  confirm its great superiorities in inference efficiency and model compactness.

\bmhead{Acknowledgments}
\label{ack}
This work was supported by National Key R\&D Program of China (No.2022ZD0118201), the National Science Fund for Distinguished Young Scholars (No.62025603), the National Natural Science Foundation of China (No. U21B2037, No. U22B2051, No. 62176222, No. 62176223, No. 62176226, No. 62072386, No. 62072387, No. 62072389, No. 62002305 and No. 62272401), and the Natural Science Foundation of Fujian Province of China (No.2021J01002,  No.2022J06001) and the China Fundamental Research Funds for the Central Universities (Grant No. 20720220068).

\section*{Declarations} 

\begin{itemize} 
\item Funding:
This work was supported by the National Science Fund for Distinguished Young Scholars (No.62025603), the National Natural Science Foundation of China (No. U21B2037, No. 62176222, No. 62176223, No. 62176226, No. 62072386, No. 62072387, No. 62072389, and No. 62002305), Guangdong Basic and Applied Basic Research Foundation (No.2019B1515120049), and the Natural Science Foundation of Fujian Province of China (No.2021J01002).
\item Conflict of interest/Competing interests:
Yongjian Wu is currently the expert researcher and the director of the Youtu Lab, Tencent Co., Ltd. The remaining  authors have no relevant financial or non-financial interests to disclose.
\item Ethics approval:
The authors have no relevant ethics approval  to disclose.
\item Consent to participate:
All authors agreed to participate in this work and made clear contributions.
\item Consent for publication:
 All authors agreed with the content and that all gave explicit consent to submit and that they obtained consent from the responsible authorities at the institute/organization where the work has been carried out.
\item Availability of data and materials:
The datasets  analysed during the current study are available in these repositories:

RefCOCO, RefCOCO+, RefCOCOg and Referit: \url{https://github.com/lichengunc/refer}; 

Flick30K Entities: \url{https://github.com/BryanPlummer/flickr30k_entities};

CLEVR: \url{https://cs.stanford.edu/people/jcjohns/clevr/};

 CLEVR-Ref+: \url{https://www.cs.jhu.edu/~cxliu/2019/clevr-ref+.html};
 
 Visual Genome: \url{ https://visualgenome.org/}.
\item Code availability:
Our code will be released after acceptance.
\item Authors' contributions:
All authors contributed to the study conception and design. Material preparation, data collection and analysis were performed by Gen Luo, Yiyi Zhou, Xiaoshuai Sun and Yue Gao. The first draft of the manuscript was written by Gen Luo and all authors commented on previous versions of the manuscript. All authors read and approved the final manuscript. More detailed contributions of each author are listed  bellow:

Conceptualization: Gen Luo, YiyiZhou and Yongjian Wu; Methodology: Gen Luo, Yiyi Zhou, Xiaoshuai Sun and Yue Gao; Writing - original draft preparation: Gen Luo; Writing - review and editing: Yiyi Zhou, Xiaoshuai Sun, Yongjian Wu, Yue Gao and Rongrong Ji; Supervision: Yiyi Zhou and Rongrong Ji.
\end{itemize}

\bibliography{sn-bibliography}


\end{document}